\documentclass[lettersize,journal]{IEEEtran}
\usepackage{amsmath,amsfonts}
\usepackage{algorithmicx}
\usepackage{algorithm}
\usepackage{array}
\usepackage[caption=false,font=normalsize,labelfont=sf,textfont=sf]{subfig}
\usepackage{textcomp}
\usepackage{stfloats}
\usepackage{url}
\usepackage{verbatim}
\usepackage{graphicx}
\usepackage{cite}
\usepackage{times}
\usepackage{epsfig}
\usepackage{amssymb}

\usepackage{booktabs}
\usepackage{xcolor}
\usepackage{multirow}
\usepackage{bbding}
\usepackage{makecell}
\usepackage{colortbl} 
\usepackage{float}
\usepackage{blindtext}
\usepackage{arydshln}
\usepackage[accsupp]{axessibility}

\usepackage[noend]{algpseudocode}

\newcommand{\eg}[0]{\emph{e.g.}}

\newcommand{\tabincell}[2]{\begin{tabular}{@{}#1@{}}#2\end{tabular}}

\usepackage{etoolbox}
\makeatletter
\patchcmd{\@makecaption}
  {\scshape}
  {}
  {}
  {}
\makeatother

\begin{document}

\font\myfont=ptmr7t at 23.36pt
\title{\myfont SLCA++: Unleash the Power of Sequential Fine-tuning \\ for Continual Learning with Pre-training}

\author{Gengwei Zhang, Liyuan Wang, Guoliang Kang, Ling Chen, Yunchao Wei,~\IEEEmembership{Member,~IEEE}
\thanks{\textit{\\
    Gengwei Zhang, Ling Chen are with Australian Artificial Intelligence Institute, University of Technology Sydney, Sydney, NSW, Australia; 
    Liyuan Wang is with Tsinghua University, Beijing, China; 
    Guoliang Kang is with Beihang University, Beijing, China; 
    Yunchao Wei is with Institute of Information Science, Beijing Jiaotong University, Beijing, China (email: \{zgwdavid, kgl.prml, wychao1987\}@gmail.com; wly19@mail.tsinghua.org.cn;
  ling.chen@uts.edu.au). 
    Gengwei Zhang and Liyuan Wang are co-first authors.
    Corresponding authors: Yunchao Wei.
    }} 
    
}

\markboth{Journal of \LaTeX\ Class Files,~Vol.~14, No.~8, August~2021}%
{Shell \MakeLowercase{\textit{et al.}}: A Sample Article Using IEEEtran.cls for IEEE Journals}


\maketitle

\begin{abstract}
   In recent years, continual learning with pre-training (CLPT) has received widespread interest, instead of its traditional focus of training from scratch. The use of strong pre-trained models (PTMs) can greatly facilitate knowledge transfer and alleviate catastrophic forgetting, but also suffers from progressive overfitting of pre-trained knowledge into specific downstream tasks. A majority of current efforts often keep the PTMs frozen and incorporate task-specific prompts to instruct representation learning, coupled with a prompt selection process for inference. However, due to the limited capacity of prompt parameters, this strategy demonstrates only sub-optimal performance in continual learning. In comparison, tuning all parameters of PTMs often provides the greatest potential for representation learning, making sequential fine-tuning (Seq FT) a fundamental baseline that has been overlooked in CLPT. To this end, we present an in-depth analysis of the progressive overfitting problem from the lens of Seq FT. Considering that the overly fast representation learning and the biased classification layer constitute this particular problem, we introduce the advanced Slow Learner with Classifier Alignment (SLCA++) framework to unleash the power of Seq FT, serving as a strong baseline approach for CLPT. Our approach involves a Slow Learner (SL) to selectively reduce the learning rate of backbone parameters, and a Classifier Alignment (CA) to align the disjoint classification layers in a post-hoc fashion. We further enhance the efficacy of SL with a symmetric cross-entropy loss (SCE), as well as employ a parameter-efficient strategy to implement Seq FT with SLCA++. Across a variety of continual learning scenarios, including class-incremental learning on general datasets like CIFAR-100 and ImageNet-R, fine-grained datasets like CUB-200 and Cars-196, and domain-incremental learning on DomainNet, our approach provides substantial improvements and outperforms state-of-the-art methods by a large margin. Our code is available at: \url{https://github.com/GengDavid/SLCA}.
\end{abstract}


\section{Introduction}
\label{sec:intro}
The purpose of continual learning (CL) is to learn contents that appear in sequence as if they were observed simultaneously. Previous efforts are mainly based on the premise of learning from scratch, attempting to mitigate catastrophic forgetting \cite{mcclelland1995there} of previously-learned knowledge when acquiring new information. 
However, the success of large-scale pre-training has revolutionized the learning paradigm of deep neural networks. The use of pre-training brings positive knowledge transfer and robustness to catastrophic forgetting for continual learning of specific downstream tasks \cite{wang2023comprehensive}, which tend to be more significant as the scale of pre-training increases \cite{ramasesh2021effect,mehta2021empirical}.
Therefore, continual learning with pre-training (CLPT) turns out to be an emerging direction and receives growing attention.

\begin{figure}[t]
    \centering
    \includegraphics[width=1.0\linewidth]{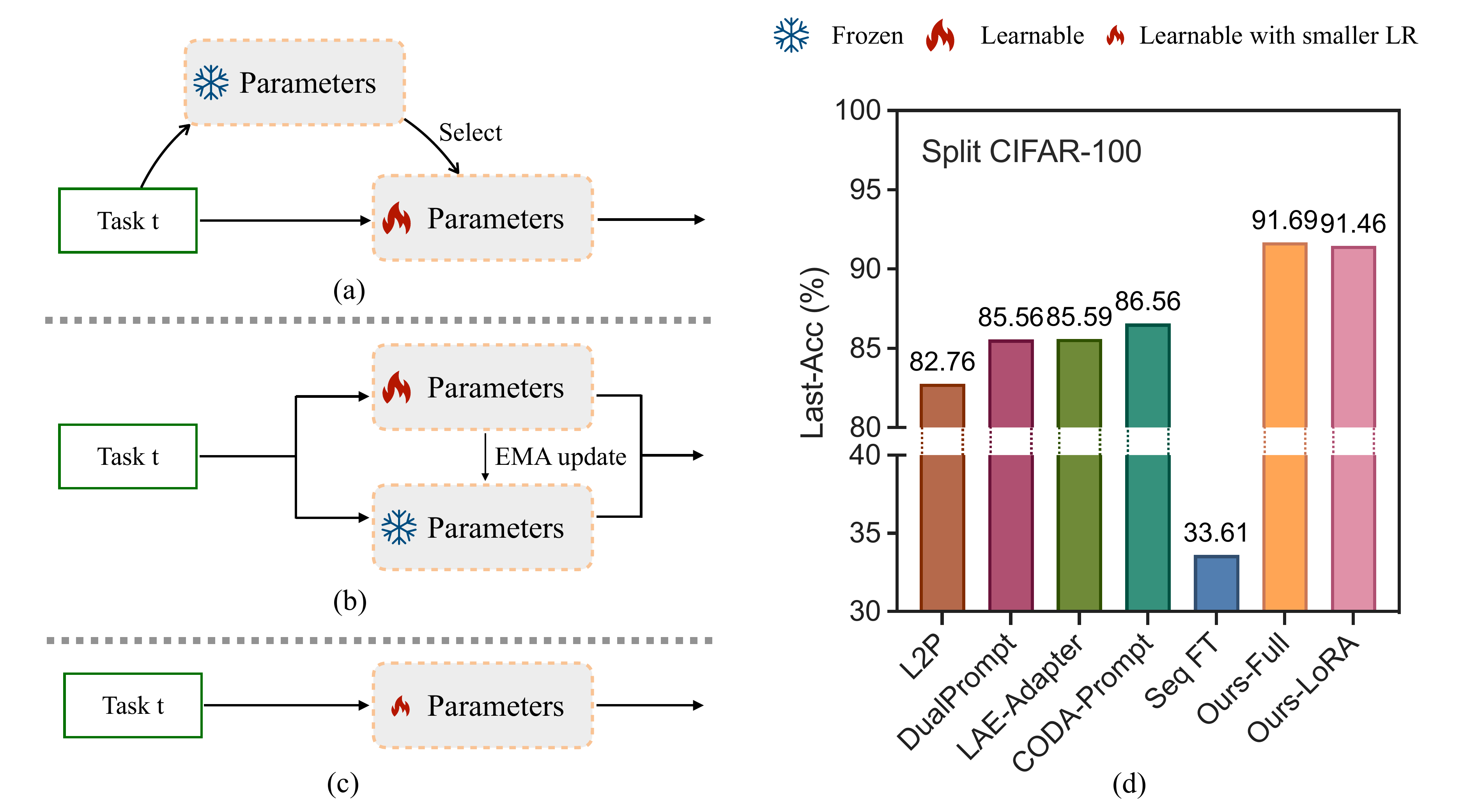}
    \caption{Comparison of recent methods and our proposal. (a) Prompt-based methods (L2P \cite{wang2022l2p}, DualPrompt \cite{wang2022dualprompt}, CODA-Prompt \cite{smith2023coda}, etc.) often construct and select appropriate prompt parameters for each task while keeping the backbone frozen.
    (b) LAE~\cite{gao2023unified} employs a momentum copy of additional parameters to stabilize their updates in CL. 
    (c) Our proposal is implemented with the simplest baseline, i.e., sequential fine-tuning (Seq FT), unleashing its power with Slow Learner (SL) and Classifier Alignment (CA).
    (d) Evaluation of continual learning performance on Split CIFAR-100 with ImageNet-21K supervised pre-training.} 
    \label{img21k_sup}
\end{figure}

CLPT poses a particular challenge that the pre-trained knowledge should be adapted to each incremental task while maintaining generalizability for future tasks.
In this regard, recent prompt-based methods \cite{wang2022l2p,wang2022dualprompt,smith2023coda} propose to freeze the pre-trained backbone (i.e., the pre-trained knowledge carried therein) and introduce a few prompt parameters to instruct representation learning (see Fig~\ref{img21k_sup}), which often involve construction and selection of appropriate prompt parameters for specific tasks. Additionally, a concurrent work~\cite{gao2023unified} devises a unified CLPT framework of parameter-efficient fine-tuning (PEFT) techniques, such as prompt \cite{lester2021power}, LoRA \cite{hu2021LoRA}, and adapter \cite{houlsby2019parameter}, which employs the exponential moving average (EMA) of additional parameters to stabilize their updates. Despite some promising results, these methods remain clearly sub-optimal in CL of specific tasks, mainly due to the limited capacity of the additional parameters  \cite{wang2024hierarchical,wang2024hide}. 
In fact, PTMs tend to have stronger adaptability as more parameters can be adjusted, where tuning all parameters often provides the largest potential for representation learning. As a result, sequential fine-tuning (Seq FT) may serve as a fundamental baseline overlooked in CLPT, although it typically represents the worst case of continual learning from scratch. Accordingly, we retreat to a simple yet important question: What is the sufficient inductive bias to make Seq FT robust in CLPT?

To this end, we perform an in-depth analysis of CLPT in terms of Seq FT. First, for a range of representative continual learning methods based on Seq FT, using a maximum stable learning rate~\cite{smith2019super} for all parameters leads to extremely inferior performance. This problem is largely due to the \emph{progressive overfitting} of pre-trained knowledge into specific downstream tasks, thus gradually losing its generalizability and stability. We further validate this insight from the perspective of gradients and arouse our solution. We observe that this problem within the representation layer (i.e., the backbone) can be almost avoided by selectively reducing the learning rate, which is sufficient to balance pre-trained knowledge and task-specific knowledge.
On the basis of a desirable representation layer, we identify that the classifier (i.e., the output layer in general) suffers from remarkable bias between tasks or classes, which require further alignment after continual learning. In this regard, we propose Slow Learner with Classifier Alignment (SLCA++), a strong baseline approach for CLPT. The former refers to the selectively reduced learning rate, while the latter employs class-wise distributions to rectify the classifier in a post-hoc fashion. To improve both efficacy and efficiency, we incorporates a SCE loss and a hybrid parameter-efficient fine-tuning strategy called Hybrid Slow Learner (Hybrid-SL). SCE furthur enhances the effect of slow learner while Hybrid-SL utilizes slow learner to tune all parameters by incorporating the idea of low-rank adaptation.

Across a variety of continual learning benchmarks under supervised and self-supervised pre-training, our approach provides substantial improvements in CLPT, and significantly fills the gap of current progress from the joint training performance. Specifically, our approach consistently improves the regular Seq FT by more than \textbf{45\%} on Split CIFAR-100, Split ImageNet-R, Split CUB-200, and Split Cars-196, thus outperforming the SOTA methods by a large margin. On Split CIFAR-100 and Split ImageNet-R, the performance gap is shorten to less than \textbf{2\%} for supervised pre-training and less than \textbf{4\%} for self-supervised pre-training. 
In particular, we find that other competitors suffer from remarkable performance drop under the more realistic self-supervised pre-training and more challenging fine-grained datasets, which makes our approach more advantageous and also points the way to subsequent work on CLPT.

Note that this work is built on a preliminary version presented at ICCV 2023 \cite{zhang2023slca}, which introduced the idea of Slow Learner with Classifier Alignment (SLCA) and revealed the potential of Seq FT for all parameters in CLPT. The current version improves upon the previous one with more in-depth analysis, methodological enhancements and more extensive experiments. First, we provide a more in-depth analysis for CLPT, and explain its particular challenge from a gradient perspective. This explanation further reveals the intrinsic mechanisms of SLCA and helps to understand its implications for CLPT. On the methodology aspect, we devise a parameter-efficient fine-tuning technique to implement SLCA, named as Hybrid Slow Learner, and the SCE loss to improve the performance, collectively referred to as SLCA++. As a result, it reduces the learnable parameters from 85.80M to 0.64M, with comparable or even better performance than before (\textit{e.g.}, improving the performance on Split Cars-196 from 67.73\% to 73.97\%). In the experimental section, we include extensive explorations of parameter-efficient fine-tuning for SLCA, evaluate domain-incremental learning, as well as compare our method with more recent baselines, so as to justify the generality of SLCA++.

Our contributions include four aspects: (1) We present an in-depth analysis of CLPT, and demonstrate that the progressive overfitting problem is the key challenge for representative continual learning methods based on Seq FT. A principle explanation from the gradient perspective based on the progressive overfitting problem is included. 
(2) We propose a simple but effective approach named SLCA++ to unleash the power of sequential fine-tuning in CLPT, which clearly outperforms state-of-the-art competitors, serving as a strong baseline to re-evaluate the current progress and technical route.
(3) We devise two strategies to further improve the efficacy and efficiency of this strong baseline, making it more applicable to continual learning scenarios.
(4) Our results further identify critical factors and promising directions for CLPT, such as pre-training paradigm and downstream granularity, so as to facilitate subsequent research.

\section{Related Work}
\label{sec:related}
\textbf{Continual Learning}. Traditional works on continual learning mainly focus on sequential training of deep neural network(s) from scratch, ensuring effective learning of new tasks without catastrophic forgetting of old tasks. 
Representative strategies include regularization-based methods \cite{kirkpatrick2017overcoming,aljundi2018memory,zenke2017continual,li2017learning,wang2021afec,dhar2019learning}, which preserve the old model and selectively stabilize changes in parameters or predictions; replay-based methods \cite{wu2019large,prabhu2020gdumb,buzzega2020dark,wang2021memory,wang2021ordisco,wang2021triple}, which approximate and recover the previously-learned data distributions; architecture-based methods \cite{serra2018overcoming,rusu2016progressive,wang2022coscl,yang2022continual}, which allocate dedicated parameter sub-spaces for each task. In addition, current advances in continual learning concentrate on the field of computer vision and gradually extend to the field of natural language processing \cite{wang2023comprehensive}.

\textbf{Continual Learning with Pre-training}. In recent years, the benefits of pre-training for continual learning have been increasingly explored.
For example, the representations obtained from supervised pre-training have been shown to facilitate not only knowledge transfer but also robustness to catastrophic forgetting in continual learning of specific downstream tasks \cite{ramasesh2021effect,mehta2021empirical,zhu2023ctp}.
Also, learning a large number of base classes in the initial training phase allows the model to learn new classes with only minor adaptations \cite{wu2022class}.
Inspired by the techniques of knowledge transfer in natural language processing, L2P \cite{wang2022l2p} employs an additional set of learnable parameters called ``prompts'' that dynamically instruct the representation layer for learning incremental tasks. DualPrompt \cite{wang2022dualprompt} extends this idea by attaching complementary prompts to the representation layer for learning task-invariant and task-specific instructions. Both L2P and DualPrompt require a prompt selection phase before adaptation. CODA-Prompt~\cite{smith2023coda} improves the utilization of prompts by an attention operation instead of the hard selection. Recently, LAE~\cite{gao2023unified} introduces a unified framework that includes different forms of parameter sets such as LoRA \cite{hu2021LoRA}, adapter \cite{houlsby2019parameter} besides prompt. These methods are reported to be far superior to representative continual learning methods based on sequential training (refer to as Seq FT in the context of continaul learning with pre-training), which potentially challenges the current paradigm of using pre-trained knowledge in computer vision.

\textbf{Parameter-efficient Fine-tuning}. The \textit{pre-training then fine-tuning} is the most adopted paradigm in transfer learning, where full fine-tuning guarantees a powerful performance. As the growth of pre-trained model size, parameter-efficient fine-tuning is introduced for improving efficiency. Partial tuning~\cite{he2022masked} is a straightforward way that freezes most backbone parts and only updates a small portion of parameters. Another idea is to keep the whole backbone fixed and introduces extra parameters like side-tuning~\cite{zhang2020side},  adapter~\cite{rebuffi2017learning,houlsby2019parameter} and low-rank adaptation~\cite{hu2021LoRA}. Recently, prompt-tuning~\cite{lester2021power} is introduced in language processing for fast adaptation and also introduces to computer vision~\cite{jia2022visual}. However, visual prompt tuning~\cite{jia2022visual} is find to struggle with adapting self-supervised pre-training. 

\textbf{Self-supervised Pre-training}. Since the large amount of training samples required to construct strong PTMs are typically unlabeled and may also arrive incrementally, self-supervised pre-training emerges as a more preferable choice than supervised pre-training. Several recent studies discover that continual learning in a self-supervised manner suffers from less catastrophic forgetting \cite{hu2021well,madaan2021rethinking,fini2022self}. Indeed, self-supervised paradigms have been shown to be better adapted to upstream continual learning, i.e., continual learning of generalized representations \cite{cossu2022continual}.
However, the effectiveness of self-supervised pre-training for downstream continual learning, i.e., continual learning based on a self-supervised pre-trained model, remains to be investigated.

\section{Continual Learning with Pre-training}
\label{sec:method}
In this section, we introduce the problem formulation of continual learning with pre-training (CLPT) and perform an in-depth analysis of its particular challenge. We then present our approach based on the analysis.

\subsection{Problem Formulation}
\label{sec:method_formulation}
\textbf{CLPT Setup}. Let's consider a neural network $M_{\theta}(\cdot) = h_{\theta_{cls}}(f_{\theta_{rps}}(\cdot))$ with parameters $\theta=\{\theta_{rps}, \theta_{cls}\}$ for classification tasks, which often consists of a representation layer $f_{\theta_{rps}}(\cdot)$ that projects input images to feature representations, and a classification layer $h_{\theta_{cls}}(\cdot)$ that projects feature representations to output predictions. $\theta_{rps}$ is initialized on a pre-training dataset $D_{pt}$ in a supervised or self-supervised manner (class labels are not necessary for the latter). 
Then, $M_{\theta}$ needs to learn a sequence of incremental tasks from their training sets $D_t, t=1,...,T$ and tries to perform well on their test sets.
Following previous efforts of CLPT in the field of computer vision~\cite{wang2022l2p,wang2022dualprompt}, we mainly focus on the class-incremental setting of continual learning \cite{vandeven2019three}. In details, $D_t = {\bigcup}_{c \in \mathbb{C}_t} \{(x_{c,n}, y_{c,n})\}_{n=1}^{N_c} $ introduces a set of new classes $\mathbb{C}_t$, where $N_c$ denotes the number of training samples $(x_{c,n}, y_{c,n})$ for class $c$, and all the classes ever seen are evaluated without task labels. Besides, in the experiments, we also evaluate our approach under the domain-incremental setting \cite{wang2022s}, where $D_t = \{(x_{n}, y_{n})\}_{n=1}^{N_{t}}$ and $N_{t}$ denotes the number of training samples belonging to domain $t$. Each $D_t$ introduce training samples from new domains, while the set of classes $\mathbb{C}$ is determined at the first task and then keep fixed for all domains.

\textbf{Progressive Overfitting}. 
To achieve the objective of CLPT, the network needs to (1) effectively transfer the pre-trained knowledge to each incremental task while maintaining its generalizability for future tasks, and (2) properly balance the learning plasticity of new tasks with memory stability of old tasks, also knwon as overcoming the catastrophic forgetting problem \cite{mcclelland1995there}. The most straightforward baseline for CLPT is to train the model $M_{\theta}$ sequentially on each $D_t$, with $f_{\theta_{rps}}$ and $h_{\theta_{cls}}$ updated at a similar speed (\textit{i.e.}, using a larger learning rate for fast convergence). However, due to the lack of $D_{pt}$ and $D_{1:t-1}$, the performance is severely restricted by a \emph{progressive overfitting} problem in both aspects. Specifically, the knowledge of $D_{pt}$ is largely interfered by $D_t$, as $\theta_{rps}$ is continually updated to accommodate incremental tasks while its generalizability obtained from the pre-training stage is progressively lost. Besides, the knowledge of $D_{1:t-1}$ is interfered by $D_t$, as $\theta_{cls}$ and $\theta_{rps}$ catastrophically forget the old tasks when learning new tasks.

To make a clear illustration, we follow the analysis of stability gap in literature~\cite{de2022continual} and express the progressive overfitting problem from a gradient perspective:
\begin{equation}
    \nabla M = \nabla M_{plas} + \nabla M_{stab} + \nabla M_{gen},
\end{equation}
where $\nabla M_{plas}$, $\nabla M_{stab}$ and $\nabla M_{gen}$ represent three gradient components for updating the model $M_{\theta}$ in continual learning. Specifically, $\nabla M_{plas}$ denotes the plasticity gradient, which aims to minimize the classification error on the current task $t$. $\nabla M_{stab}$ and $\nabla M_{gen}$ represent the stability and generalizability gradients for maintaining stability of old tasks and generalizability of pre-trained knowledge, respectively. Below we will discuss their respective impacts on the challenges of continual learning.

\begin{figure}[t]
    \centering
    \includegraphics[width=0.80\linewidth]{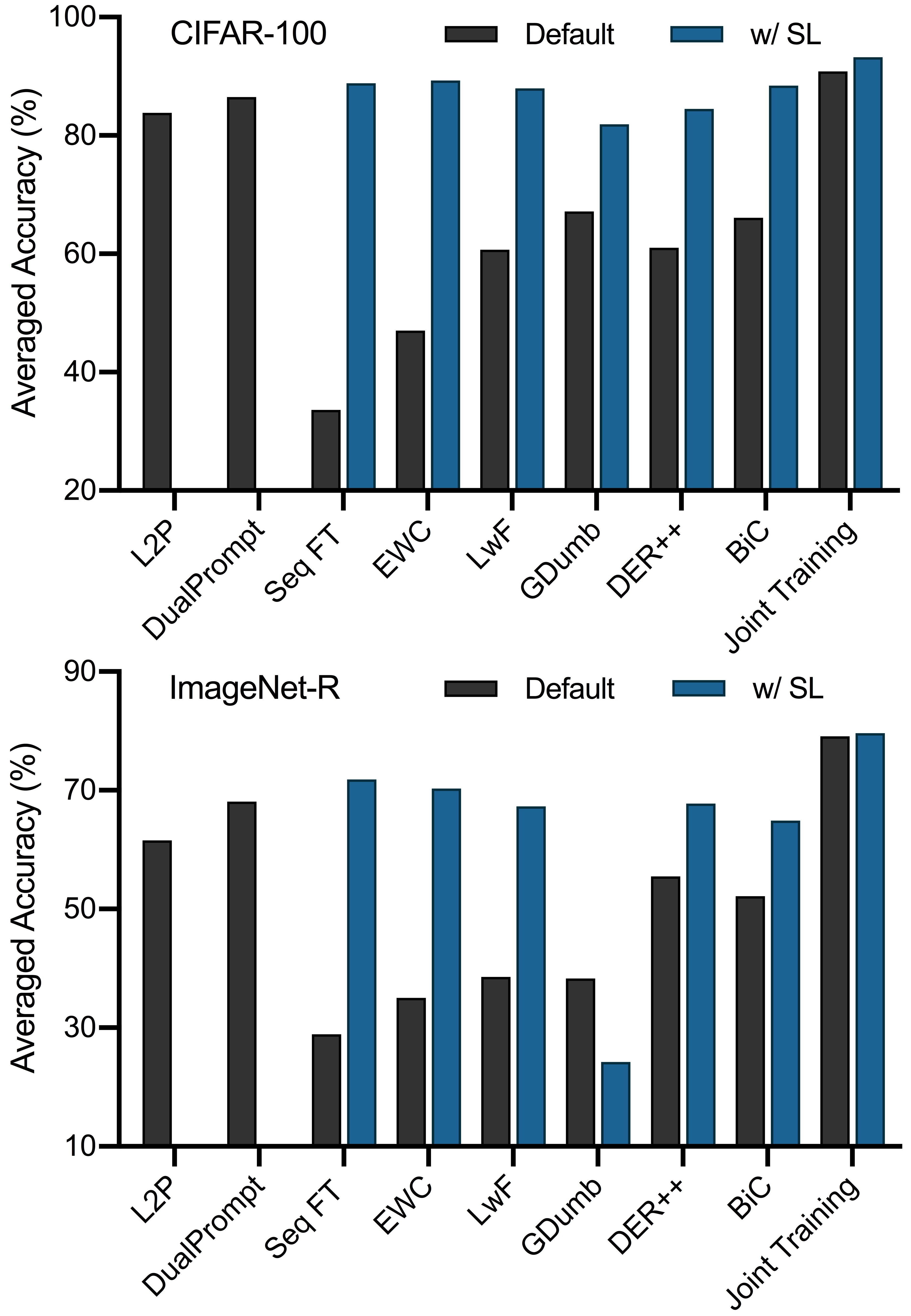}
    \caption{Slow Learner (SL) can greatly enhance the performance of sequential fine-tuning (Seq FT) in CLPT.
    Here we adopt ImageNet-21K supervised pre-training for all baselines with default performance referenced from previous efforts \cite{wang2022l2p,wang2022dualprompt},
    including prompt-based methods (L2P~\cite{wang2022l2p} and DualPrompt~\cite{wang2022dualprompt}), regularization-based methods (EWC~\cite{kirkpatrick2017overcoming} and LwF~\cite{li2017learning}), and replay-based methods (GDumb~\cite{prabhu2020gdumb}, DER++~\cite{buzzega2020dark} and BiC \cite{wu2019large}). }
    \label{img21k_sup_all}
\end{figure}

\begin{figure}[t]
    \centering
    \includegraphics[width=0.82\linewidth]{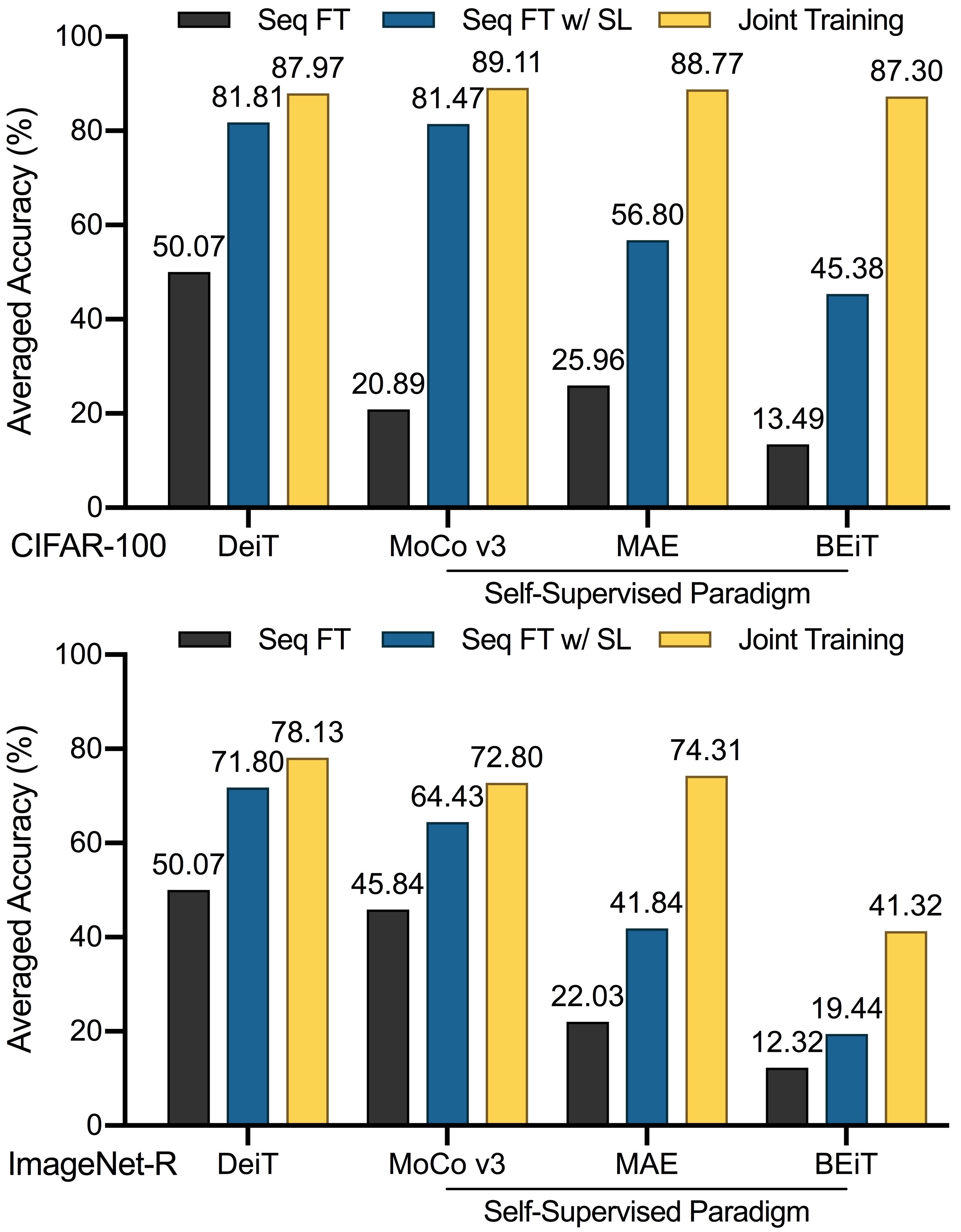}
    \caption{Comparison of pre-training paradigms on ImageNet-1K. DeiT \cite{touvron2021training} is a strong supervised method for (pre-)training vision transformer, while MoCo v3 \cite{chen2021empirical}, MAE \cite{he2022masked} and BEiT \cite{bao2021beit} are representative self-supervised methods. The pre-trained checkpoints are obtained from their official release. } 
    \label{img1k-cifar100-pt}
\end{figure}

\subsection{Slow Learner is (Almost) All You Need?} 
For continual learning from scratch, sequential fine-tuning (Seq FT) represents the worst-case performance in general. This is because $\| \nabla M_{gen} \| = 0$ and $\| \nabla M_{stab} \|$ is often close to $0$, and $\nabla M_{plas}$ dominates the training gradients and results in catastrophic forgetting. When $M_{\theta}$ is pre-trained with $D_{pt}$, the large-scale nature of $D_{pt}$ \textit{implicitly} provides $M_{\theta}$ a large $\nabla M_{gen}$, making it success when fine-tuning on a wide range of separate downstream tasks. Therefore, a question is natually raised: \textit{whether the pre-trained weights can also implicitly supply us a proper $\nabla M_{stab}$ like it does in providing $\nabla M_{gen}$?} 
In previous efforts on CLPT \cite{wang2022l2p,wang2022dualprompt}, the answer is possibly ``No'' since Seq FT still performs poorly in their implementation. By careful dissecting, we find that they use a relatively large learning rate (\textit{e.g.,} 0.005, which can make joint-training converges well on all datasets) for both $\theta_{rps}$ and $\theta_{cls}$. We conjecture that $\nabla M_{stab}$ is obscured in this case by a large $\| \nabla M_{plas} \|$, which is independent to $\nabla M_{stab}$ and $\nabla M_{gen}$ and can be explicitly modified by adjusting the learning rate. 
Surprisingly, when using a much smaller learning rate (0.0001) for $\theta_{rps}$ and a slightly larger learning rate (0.01) for $\theta_{cls}$, the sequential fine-tuning (Seq FT) baseline is greatly enhanced. As shown in Fig.~\ref{img21k_sup_all}\footnote{A more extensive analysis of the impact of learning rates is presented in Appendix~\ref{sec:append_ana}.}, Seq FT is improved by more than 40\% for challenging continual learning benchmarks such as Split CIFAR-100 and Split ImageNet-R, respectively. Besides, we find that such a simple change also makes Seq FT clearly outperform the recent prompt-based methods such as L2P \cite{wang2022l2p} and DualPrompt \cite{wang2022dualprompt}.
These prompt-based methods~\cite{wang2022l2p,wang2022dualprompt,smith2023coda} fixes the representation layer and employes an additional set of learnable parameters $\theta_{add}$ to instruct the pre-trained model. From the gradient optimization perspective, although it makes $\| \nabla M_{gen} \| \approx \infty$ for the fixed representation layer, the newly added parameters $\theta_{add}$ would suffer from even severe progressive overfitting problem with random initialization. 

We call the simple but remarkably effective strategy ``Slow Learner (SL)'', corresponding to slowing down the updating speed, \textit{i.e.}, reducing $\| \nabla M_{plas} \|$ of the representation layer. It reveals that the use of pre-training indeed implicitly introduces $\nabla M_{stab}$ and $\nabla M_{gen}$ to regularize modifications of the network, but was obscured and thus overlooked by previous efforts. 
It is noteworthy that although the use of different learning rates for different network layers has been explored in transfer learning \cite{guo2019spottune,zhang2020revisiting,he2019rethinking}, its \emph{specific benefits} for continual learning has not been discovered yet. In particular, while the upper bound performance (\textit{i.e.}, joint training) is similar or marginally improved by the SL, the performance gap between continual learning and joint training is greatly filled (\textit{e.g.}, only 4.36\% on Split CIFAR-100 and 7.80\% on Split ImageNet-R for Seq FT w/ SL). 

Besides, we also apply SL to several representative continual learning methods that explicitly introduce $\nabla M_{stab}$ by either regularization loss~\cite{kirkpatrick2017overcoming,li2017learning} or data replay~\cite{buzzega2020dark,prabhu2020gdumb,wu2019large} beyond Seq FT. As shown in Fig.~\ref{img21k_sup_all}, combining SL with these methods also achieves substantial improvements, demonstrating that the benefits of $\nabla M_{stab}$ provided by pre-training remain significant even with anti-forgetting techniques. On the other hand, these methods are based on the assumption of training from scratch, and their performance is only comparable to Seq FT w/ SL. The results indicate that $\nabla M_{stab}$ provided by pre-training is already sufficient for maintaining stability, and introducing additional $\nabla M_{stab}$ from old tasks would cause unnecessary suppression of $\nabla M_{gen}$ and $\nabla M_{plas}$, thereby limiting the final performance.

\textbf{Effect of Pre-training Paradigm}. Given the effectiveness of SL in supervised pre-training, we are curious about whether this useful property also generalize well to self-supervised pre-training, which avoids any explicit labels during the pre-training phase. 
Considering architectural consistency with previous efforts of CLPT \cite{wang2022l2p,wang2022dualprompt}, we select representative self-supervised methods (\textit{i.e.,} MoCo v3 \cite{chen2021empirical}, MAE \cite{he2022masked} and BEiT \cite{bao2021beit}) that release checkpoints on ViT-B/16 in our experiments. We also compare with DeiT \cite{touvron2021training}, a strong supervised method for (pre-)training vision transformer on ImageNet-1K dataset.
As shown in Fig.~\ref{img1k-cifar100-pt}, self-supervised pre-training, while more realistic regarding labeling requirements and upstream continual learning, typically results in a larger performance gap between Seq FT and joint training than supervised pre-training. 
Similarly, the use of SL can effectively reduce this gap, suggesting that self-supervised pre-training also provides $\nabla M_{stab}$ similarly as supervised pre-training. 

\begin{figure}[t]
    \centering
    \includegraphics[width=0.90\linewidth]{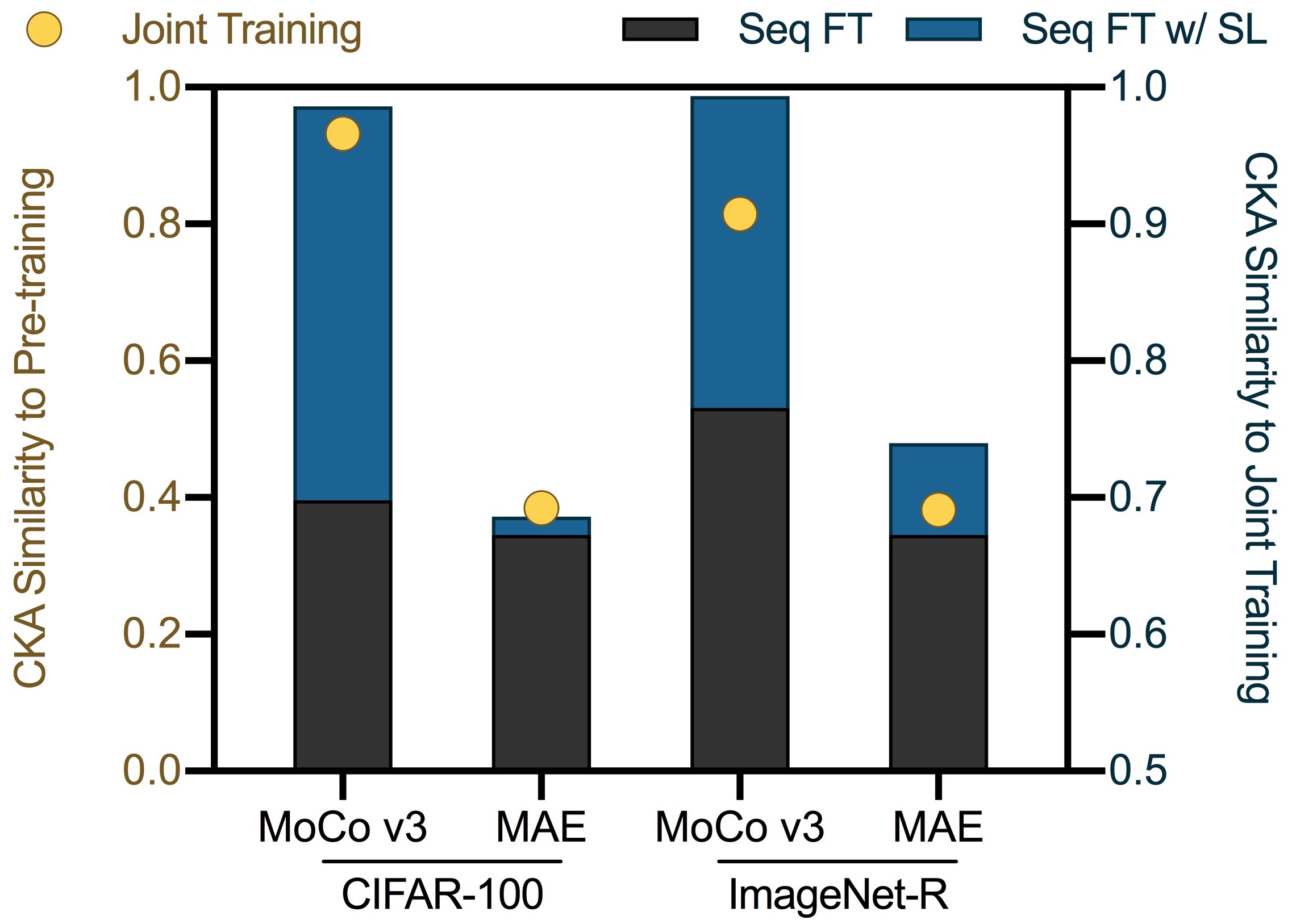}
    \caption{Similarity of the pre-trained representations (1) before and after joint training (left Y-axis, yellow dot), and (2) after joint training and after continual learning (right Y-axis, column).
    We adopt Centered Kernel Alignment (CKA) \cite{kornblith2019similarity} as the similarity metric. \emph{Best viewed in color.}
    }
    \label{ssl_cka_jt_seq}
\end{figure}

Interestingly, the performance of Seq FT w/ SL for MoCo v3 \cite{chen2021empirical} far exceeds that of the more recent MAE \cite{he2022masked}, although their joint training performance is comparable. Meanwhile, the use of SL allows MoCo v3 \cite{chen2021empirical} to learn representation much closer to that of the joint training (Fig.~\ref{ssl_cka_jt_seq}, right Y-axis). In contrast, the continually learned representation with MAE pre-training are neither close to that of the joint training nor to the initial pre-training. From the gradient perspective, a larger $\| \nabla M_{plas} \|$ is required for the MAE pre-training, while the overall strength $\| \nabla M_{stab} \|$ provided is very limited, making the change of representation behave similarly to that of training from scratch in continual learning. 
In short, the above analysis suggests a new direction for designing the paradigms of self-supervised pre-training, \textit{i.e.}, how to benefit downstream continual learning and combine the advantages of SL.

\begin{figure*}[t]
    \centering
    \includegraphics[width=1\linewidth]{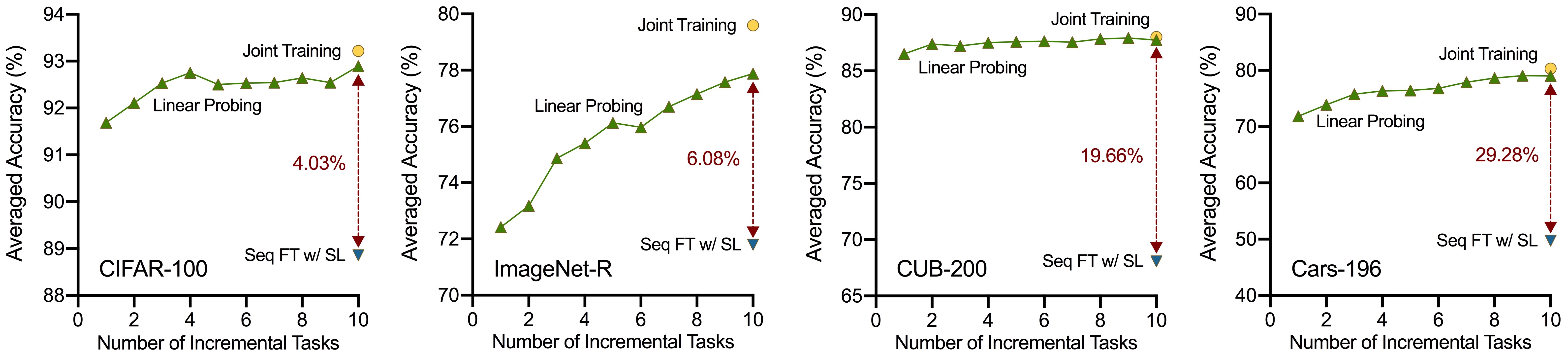}
    \caption{Linear probing results of the Slow Learner. All experiments are based on ImageNet-21K supervised pre-training. We report the averaged accuracy of all classes in the corresponding benchmark dataset (\textit{e.g.}, a total of 100 classes in CIFAR-100 dataset). The dark red arrow represents the performance gap caused by a sub-optimal classification layer.
    }
    \label{img21k-cifar100-probe}
\end{figure*}

\textbf{Evaluation of Representation}. 
Moreover, how does SL contribute to the representation learning over incremental tasks? What accounts for the remaining performance gap? 
To answer these questions, here we perform a linear probing experiment \cite{he2022masked} to evaluate the performance of the representation layer. Specifically, after learning each incremental task (\textit{e.g.}, 10 classes per task for 10 tasks in Split CIFAR-100) via Seq FT w/ SL, we fix the representation layer and employ an extra classification layer, called a linear probe, to learn all classes of the corresponding benchmark dataset (\textit{e.g.}, a total of 100 classes in CIFAR-100 dataset). The performance of these linear probes is presented in Fig.~\ref{img21k-cifar100-probe}, which tends to grow with learning more tasks, indicating that the representation layer is accumulating knowledge for better adaptation. After learning all incremental tasks, it can be clearly seen that using the continually learned representation layer to jointly train an extra classifier for all classes can almost reach the joint training performance of the entire model, and far outperform its counterpart with a continually learned classifier (\textit{i.e.}, Seq FT w/ SL in Fig.~\ref{img21k-cifar100-probe}). Therefore, the proposed SL can almost address the problem of the representation layer, yet the classification layer remains sub-optimal. In particular, the problem of classification layer becomes more severe for fine-grained continual learning benchmarks such as Split CUB-200 and Split Cars-196. This phenomenon is mainly caused by the disjoint optimization of the classifier $h_t$ in each task, \textit{i.e.,} only the parameters $\theta_{cls,t}\in \mathbb{R}^{d\times C_t}$ ($C_t$ is the number of classes in $\mathbb{C}_t$) that corresponds to the classes of the current task are trained together with the representation layer $f_{\theta_{rep}}$.

\begin{figure}[t]
    \centering
    \includegraphics[width=1\linewidth]{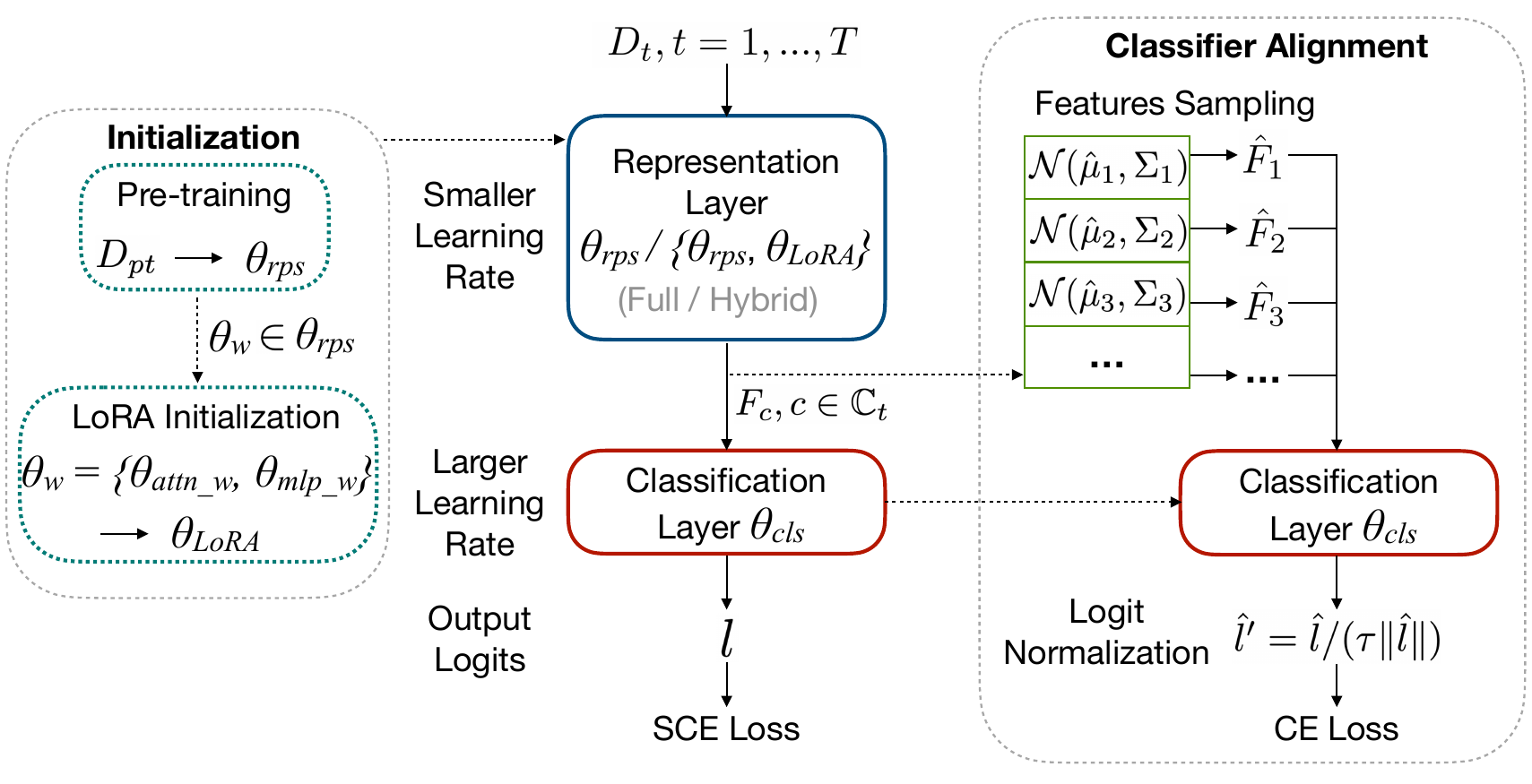}
    \caption{Illustration of the proposed SLCA++ approach for CLPT.  }
    \label{SLCA}
\end{figure}

\subsection{Slow Learner with Classifier Alignment}
To further improve the classification layer, we propose to save statistics of each class during the continual learning progress and then align all classifiers in a \textit{post-hoc} fashion (see Fig.~\ref{SLCA} and Algorithm~\ref{pesudocode}), called Classifier Alignment (CA). Specifically, after learning each task, we collect feature representations $F_{c}=\{r_{c,1}, ... , r_{c,N_c}\}$ for each class $c \in \mathbb{C}_t$ within the task, where $r_{c,n} = f_{\theta_{rps}}(x_{c,n})$ and $N_c$ denotes its amount. 
Instead of saving the extracted features $F_{c}$ of training samples, CA preserves their mean $\mu_c \in \mathbb{R}^{d}$ and covariance $\Sigma_c \in \mathbb{R}^{d \times d}$ for each class $c$ ($d$ denotes the feature dimension) for memory efficiency. 

Whenever the model needs to be evaluated, the classification layers are further aligned as follows.
Given the preserved mean $\mu_c$ and covariance $\Sigma_c$ for each classes, we model the feature distribution as a Gaussian $\mathcal{N}(\mu_c, \Sigma_c)$, since the use of pre-training provides well-distributed representations and each class tends to be single-peaked. 
Considering the possible semantic drift~\cite{yu2020semantic} problem, where the feature distribution after learning old tasks may not reflect the feature distribution at inference time, we introduce a slight modification to the feature mean $\mu_c$ in CA. Specifically, according to the research in open-set recognition~\cite{chen2021adversarial,dhamija2018reducing}, optimizing softmax cross-entropy loss within a finite category space would make unknown samples to have lower feature magnitude. In continual learning, training on classes of current task can lead to decrease of feature magnitude of old classes. 
Therefore, we scale down each feature mean $\hat{\mu}_c=\lambda_t \mu_c$ based on the learning progress with scaling factor $\lambda_t = \frac{1}{1+\eta*(T-t)}$, where $\eta$ controls the degree of scaling magnitude, which is set to 0.02 in all experiments, thus $\lambda_t$ is dynamically determined by the incremental progress, and $t$ is the the task identity that class $c$ belongs to. 

Next, we sample generated features $\hat{F}_c=\{\hat{r}_{c,1}, ... , \hat{r}_{c,S_c}\}$ from the distribution $\mathcal{N}(\hat{\mu}_c, \Sigma_c)$ of each class $c \in \mathbb{C}_{1:T}$, where $S_c$ is the amount of generated features for each class ($S_{c}$ is set to 256 in all experiments), and the number of tasks ever seen $T$ can be any positive integer without being known in advance. 
The generated features $\hat{F}_{1:T} = \{\hat{F}_1,...,\hat{F}_{C_{1:T}}\}$ ($C_{1:T}$ is the total number of classes in $\mathbb{C}_{1:T}$) is feed to the classification layer $h_{\theta_{cls}}$ as input, and a widely-used cross-entropy loss is adopted to furthur optimze the classification layer.

\begin{algorithm}[ht]
\caption{\small Slow Learner with Classifier Alignment (SLCA++) }
\label{pesudocode}
{\bf Inputs:} Pre-training dataset $D_{pt}$; training dataset $D_t$ for task $t=1,...,T$; network $M_{\theta}(\cdot) = h_{\theta_{cls}}(f_{\theta_{rps}}(\cdot))$ with parameters $\theta=\{\theta_{rps}, \theta_{cls}\}$; LoRA parameters $\theta_{LoRA}$; learning rates $\alpha$ for $\theta_{rps}$ and $\beta$ for $\theta_{cls}$ ($\alpha < \beta$); temperature hyperparameter $\tau$.
\begin{algorithmic}[1]
\State \textit{\# Initialization.} 
\State Initialize $\theta_{rps}$ by pre-training on $D_{pt}$; initialize $\theta_{LoRA}$ according to $\theta_{rep}$; initialize $\theta_{cls}$ randomly. 
\If{full version}
\State $\theta_{rep} := \theta_{rep}$
\ElsIf{hybrid version}
\State $\theta_{rep} := \{\theta_{rep}, \theta_{LoRA}\}$ with $\theta_w$ not activated 
\EndIf
\State \textbf{end if}
\State \textit{\# Sequential tasks.} 
\For{task $t=1,...,T$} 
 \State \textit{\# different learning rates $\alpha$, $\beta$ for $\theta_{rep}$ and $\theta_{cls}$.}
 \While{not converged} 
 \State 
 Train $M_{\theta}$ on $D_t$ with SCE loss in Eqn.~\ref{eqn_sce}.
 \EndWhile
 \State \textbf{end while}
  \State Collect $F_c=\{r_{c,1}, ... , r_{c,N_c}\}$ for $c \in \mathbb{C}_t$.
  \State Save mean $\mu_c$ and covariance $\Sigma_c$ of $F_c$ for $c \in \mathbb{C}_t$.
\EndFor
\State \textbf{end for}
  \State \textit{\# Classifier alignment.}
  \State Sample $\hat{F}_c$ from $\mathcal{N}(\hat{\mu}_c, \Sigma_c)$ for $c \in \mathbb{C}_{1:T}$.
 \While{not converged} 
 \State Compute logit $\hat{l}$ and its magnitude $\| \hat{l} \|$. 
 \State {Train $h_{\theta_{cls}}$ with normalized logit in Eqn.~\ref{eqn_modifed_ce}.}
 \EndWhile
 \State \textbf{end while}
\end{algorithmic}
\end{algorithm}

However, a prolonged training of the classification layer can lead to an overconfidence issue, which potentially impairs generalizability to the test set(s). To overcome this issue, we draw inspirations from out-of-distribution (OOD) detection \cite{wei2022mitigating} and normalize the magnitude of network outputs when computing the cross-entropy. 
Let $\hat{l}=h_{\theta_{cls}}(\hat{r})$ denote the logit (\textit{i.e.}, pre-softmax output) of a generated feature $\hat{r}$ and $\hat{l}\in \mathbb{R}^{C_{1:T}}$, which can be re-written as the product of two components: $\hat{l} = \| \hat{l} \| \cdot \vec{\hat{l}},$ 
where $\| \cdot \|$ denotes $L_2$-norm. Accordingly, $\| \hat{l} \| = \sqrt{\sum_{c \in C_{1:T}} \| l_{c} \|^2} $ represents the magnitude of $\hat{l}$, and $\vec{\hat{l}}$ represents its direction. Then we adopt a modified cross-entropy loss with logit normalization to perform CA:
\begin{equation}
\mathcal{L}(\theta_{cls}; \hat{r} ) = - \log \frac{e^{{\hat{l}_y} / (\tau \|\hat{l}\|)}}{\sum_{c \in \mathbb{C}_{1:T}} e^{{\hat{l}_{c}} / (\tau \|\hat{l}\|)}},
\label{eqn_modifed_ce}
\end{equation}
where $\hat{l}_y$ denotes the $y$-th element of $\hat{l}$ corresponding to the ground-truth label $y$. $\tau$ is a temperature hyperparameter. 
The intuition behind it is that normalizing $\hat{l}$ with an input-dependent constant $\tau \|\hat{l}\|$ will not change the result of prediction $\arg\!\max_{c \in \mathbb{C}_{1:T}}(l_{c})$, while forcing the magnitude $\|\hat{l}\|$ before softmax becomes $\frac{1}{\tau}$, which can make the criterion only adjust the direction $\vec{\hat{l}}$ \cite{wei2022mitigating}. Therefore, the normalization in Eqn.~\ref{eqn_modifed_ce} can alleviate the overconfidence issue in classifier alignment.
In practice, we observe that $\tau$ is not sensitive and empirically find $\tau = 0.1$ to be a reasonable choice.

\subsection{SLCA with Parameter-efficient Fine-tuning (PEFT)}
Our work reveals that tuning all parameters is clearly more advantageous than tuning a few inserted short sequence. Besides, with SLCA, a task-shared manner is effective enough to resolve the progressive overfitting problem. However, directly tuning all parameters is considered to be expensive for a large model. Recently, low-rank adaptation (LoRA)~\cite{hu2021LoRA} provides an efficient way of adjusting a low-rank decomposition of parameter-intensive layers, such as attention and feed-forward layers in the transformer, providing an opportunity for implementing SLCA in a parameter-efficient fashion.

Therefore, we introduce \textbf{Hybrid Slow Learner}, a parameter-efficient version that adjusts all network parameters in an efficient way. In particular, first considering components like bias, normalization layers and the class token, all these parameters have already satisfied the nature of ``parameter-efficient''. We directly fine-tune them sequentially for each task with slow learner. Instead, for the ``expensive'' parameters, \textit{i.e.}, weight parameters $\theta_w$ in the attention and feed-forward layers, an additional LoRA in used. 
Specifically, suppose we have a pre-trained weight matrix $W\in \mathbb{R}^{d_2 \times d_1}$, LoRA constrains the learning of the weight with a low-rank decomposition $W+ \Delta\!W=W+BA$, where $A \in \mathbb{R}^{k\times d_1}$, $B \in \mathbb{R}^{d_2 \times k}$, rank $k \ll \min(d1,d2)$ and only two low-rank matrices $A$ and $B$ are activated for updating during training. After all learning stages, $A$ and $B$ can be absorbed into $W$, making the network architecture unchanged. Similar to other parameters, the injected LoRA layer is sequentially fine-tuned with Slow Learner (SL) for all tasks in a task-shared manner.

\textbf{Initialization of LoRA}. Typically, the base matrix $A$ is initialized with random values drawn form Gaussian distribution, while the weighting matrix $B$ is initialized to zero to make $BA=0$ at the beginning of the training. According to the analysis of LoRA~\cite{hu2021LoRA}, the learning sub-space of LoRA has a stronger correlation with the space of pre-trained weight $W$ compared to a random matrix. We utilize this property and perform a singular value decomposition (SVD) on the pre-trained weight $W$ to get $W = U\Sigma_sV^{\top}$, where $U\in \mathbb{R}^{d_2\times d_2}$ and $V\in \mathbb{R}^{d_1\times d_1}$ are the left and right singular vectors of $W$ and $\Sigma_s \in \mathbb{R}^{d_2\times d_1}$ contains singular values of $W$. Although we can obtain the low-rank decomposition of $W$ by keeping top-$k$ elements in $U$, $\Sigma_s$ and $V^{\top}$, we only keep top-$k$ rows in $V^{\top}$ to initialize matrix $A$ while keeping $B$ as zero, since our goal is to find a proper initialization for the adaptation layer rather than to reconstruct $W$. In this way, we obtain a LoRA layer starting at $BA=0$ but with a better initialization state, which works surprisingly well with the proposed SL.

\subsection{Learning Objective}
\label{sec:sce}
Typically, we use standard softmax cross-entropy loss $\mathcal{L}_{CE}=-q\log p$ on categories of the current task for optimize the network, where $q$ is the one-hot label distribution, $p=\sigma(l)$ is the post-softmax logits and $\sigma(\cdot)$ is the softmax function. 
However, training with cross-entropy loss is known to be dominant by gradients of low confidences samples~\cite{mummadi2021test}. In joint fine-tuning or a single task training, it is beneficial as it accelerate the convergence of the network with hard samples. In contrast, since the network is sequentially trained in continual learning, the dominant gradients from the low confidences samples exacerbates the problem of progressive overfitting problem by suppressing $\nabla M_{stab}$ and enlarging $\nabla M_{plas}$, and thus deteriorates the final continual learning performance. Similar to the intuition of slow learner, a strategy for balancing the training speed (\textit{i.e.}, the magnitude of gradients here) is needed. Inspired by symmetric cross-entropy (SCE)~\cite{wang2019symmetric}, originally introduced for solving noisy label problem, we add a reverse cross-entropy (RCE) loss along with the CE loss as: 
\begin{equation}
    \mathcal{L}_{SCE}=\alpha\mathcal{L}_{CE}+\beta\mathcal{L}_{RCE},
    \label{eqn_sce}
\end{equation}
where the second term $\mathcal{L}_{RCE}=-p\log q$ is the ``reverse'' version of $\mathcal{L}_{CE}$ and two balancing factors $\alpha$ and $\beta$ are included. According to the previous analysis~\cite{wang2019symmetric,dobler2023robust}, SCE loss can balance the gradients between high and low confidence samples, thus benefit the optimization of continual learning.

\begin{table*}[ht]
    \centering
    \caption{Experimental results for continual learning on Split CIFAR-100 and Split ImageNet-R. IN21K-Sup: supervised pre-training on ImageNet-21K. IN1K-Self: self-supervised pre-training on ImageNet-1K with MoCo v3 \cite{chen2021empirical}.
    All other continual learning methods based on Seq FT are reproduced according to their public codes with the proposed \textbf{Slow Learner} implemented. \textcolor{gray}{SLCA++ (Full)}: our full version that directly tuning all parameters in the backbone.}
    \footnotesize{
    \begin{tabular}{c|l|c|c||c|c||c|c}
    \hline
        \multirow{2}{*}{Type}& \multirow{2}{*}{Method} & \multirow{2}{*}{\#Params} & \multirow{2}{*}{Pre-trained} & \multicolumn{2}{c||}{Split CIFAR-100} & \multicolumn{2}{c}{Split ImageNet-R} \\ \cline{5-8}
        & & & & Last-Acc (\%)& Inc-Acc (\%)& Last-Acc (\%)& Inc-Acc (\%)\\ \hline
        Upper-Bound & Joint-Training & 85.80M & IN21K-Sup & 93.22\tiny{$\pm 0.16$} & - & 80.76\tiny{$\pm 0.73$} & -  \\ 
        \hline
        \multirow{5}*{\tabincell{c}{Classical}} 
        & GDumb \cite{prabhu2020gdumb}& 85.80M & IN21K-Sup & 81.92\tiny{$\pm 0.15$} & 89.46\tiny{$\pm 0.94$} & 24.23\tiny{$\pm 0.35$} & 43.48\tiny{$\pm 0.49$}  \\ 
        & DER++ \cite{buzzega2020dark}& 85.80M & IN21K-Sup & 84.50\tiny{$\pm 1.67$} & 91.49\tiny{$\pm 0.61$} & 67.75\tiny{$\pm 0.93$} & 78.13\tiny{$\pm 1.14$} \\ 
        & BiC \cite{wu2019large}& 85.80M & IN21K-Sup & 88.45\tiny{$\pm 0.57$} & 93.37\tiny{$\pm 0.32$} & 64.89\tiny{$\pm 0.80$} & 73.66\tiny{$\pm 1.61$}  \\ 
        & EWC \cite{kirkpatrick2017overcoming}& 85.80M & IN21K-Sup & 89.30\tiny{$\pm 0.23$} & 92.31\tiny{$\pm 1.66$}  & 70.27\tiny{$\pm 1.99$} & 76.27\tiny{$\pm 2.13$}  \\ 
        & LwF \cite{li2017learning}& 85.80M & IN21K-Sup & 87.99\tiny{$\pm 0.05$} & 92.13\tiny{$\pm 1.16$}  & 67.29\tiny{$\pm 1.67$} & 74.47\tiny{$\pm 1.48$}  \\ 
        \hdashline
        \multirow{4}*{\tabincell{c}{Param-Efficient}} 
        & L2P \cite{wang2022l2p}& 0.46M & IN21K-Sup & 82.76\tiny{$\pm 1.17$} & 88.48\tiny{$\pm 0.83$} & 66.49\tiny{$\pm 0.40$} & 72.83\tiny{$\pm 0.56$} \\ 
        & DualPrompt \cite{wang2022dualprompt}& 0.49M & IN21K-Sup & 85.56\tiny{$\pm 0.33$} & 90.33\tiny{$\pm 0.33$} & 68.50\tiny{$\pm 0.52$} & 72.59\tiny{$\pm 0.24$}  \\
        & LAE (Adapter) \cite{gao2023unified} & 0.15M & IN21K-Sup & 85.59\tiny{$\pm 0.46$} & 89.96\tiny{$\pm 0.44$} & 72.66\tiny{$\pm 0.63$} & 78.91\tiny{$\pm 0.89$}  \\ 
        & CODA-Prompt \cite{smith2023coda} & 3.84M & IN21K+IN1K-Sup & 86.56\tiny{$\pm 0.77$} & 90.61\tiny{$\pm 0.36$} & 75.25\tiny{$\pm 0.56$} & 81.26\tiny{$\pm 0.76$}  \\ 
        \hdashline
        \multirow{4}*{\tabincell{c}{Ours}} 
        & \textcolor{gray}{SLCA~\cite{zhang2023slca}} & \textcolor{gray}{85.80M} & \textcolor{gray}{IN21K-Sup} & \textcolor{gray}{{91.53}\tiny{$\pm 0.28$}} & \textcolor{gray}{{94.09}\tiny{$\pm 0.87$}} & \textcolor{gray}{{77.00}\tiny{$\pm 0.33$}} & \textcolor{gray}{{81.17}\tiny{$\pm 0.64$}} \\
        & \textcolor{gray}{SLCA++ (Full)} & \textcolor{gray}{85.80M} & \textcolor{gray}{IN21K-Sup} & \textcolor{gray}{91.69\tiny{$\pm 0.15$}} & \textcolor{gray}{94.47\tiny{$\pm 0.72$}} & \textcolor{gray}{79.78\tiny{$\pm 0.16$}} & \textcolor{gray}{84.31\tiny{$\pm 0.73$}} \\
        & SL++ & 0.64M & IN21K-Sup & 89.86\tiny{$\pm 0.31$} & 92.88\tiny{$\pm 0.92$} &  76.41\tiny{$\pm 0.52$} & 82.05\tiny{$\pm 0.88$} \\ 
        & SLCA++ & 0.64M & IN21K-Sup & \textbf{91.46}\tiny{$\pm 0.18$} & \textbf{94.20}\tiny{$\pm 0.71$} & \textbf{78.09}\tiny{$\pm 0.22$} & \textbf{82.95}\tiny{$\pm 0.78$} \\
        \hline \hline
        Upper-Bound & Joint-Training & 85.80M & IN1K-Self & 89.11\tiny{$\pm 0.06$} & - & 72.80\tiny{$\pm 0.23$} & -  \\ \hline 
        \multirow{5}*{\tabincell{c}{Classical}} 
        & GDumb \cite{prabhu2020gdumb}& 85.80M & IN1K-Self & 69.72\tiny{$\pm 0.20$} & 80.95\tiny{$\pm 1.19$} & 28.24\tiny{$\pm 0.58$} & 43.64\tiny{$\pm 1.05$}  \\ 
        & DER++ \cite{buzzega2020dark}& 85.80M & IN1K-Self & 63.64\tiny{$\pm 1.30$} & 79.55\tiny{$\pm 0.87$} & 53.11\tiny{$\pm 0.44$} & 65.10\tiny{$\pm 0.91$} \\ 
        & BiC \cite{wu2019large}& 85.80M & IN1K-Self & 80.57\tiny{$\pm 0.86$} & 89.39\tiny{$\pm 0.33$} & 57.36\tiny{$\pm 2.68$} & 68.07\tiny{$\pm 0.22$} \\ 
        & EWC \cite{kirkpatrick2017overcoming}& 85.80M & IN1K-Self & 81.62\tiny{$\pm 0.34$} & 87.56\tiny{$\pm 0.97$} & 64.50\tiny{$\pm 0.36$} & 70.37\tiny{$\pm 0.41$} \\ 
        & LwF \cite{li2017learning}& 85.80M & IN1K-Self & 77.94\tiny{$\pm 1.00$} & 86.90\tiny{$\pm 0.90$} & 60.74\tiny{$\pm 0.30$} & 68.55\tiny{$\pm 0.65$} \\ 
        \hdashline
        \multirow{4}*{\tabincell{c}{Param-Efficient}} 
        & L2P \cite{wang2022l2p}& 0.46M & IN1K-Self & 68.35\tiny{$\pm 0.48$} & 79.28\tiny{$\pm 0.82$} & 51.53\tiny{$\pm 0.67$} & 63.96\tiny{$\pm 0.78$} \\ 
        & DualPrompt \cite{wang2022dualprompt}& 0.49M & IN1K-Self & 74.29\tiny{$\pm 0.64$} & 83.36\tiny{$\pm 0.61$} & 59.31\tiny{$\pm 0.77$} & 68.38\tiny{$\pm 1.06$} \\
        & LAE (Adapter) \cite{gao2023unified} & 0.15M & IN1K-Self & 74.87\tiny{$\pm 0.55$} & 83.73\tiny{$\pm 0.36$} & 62.81\tiny{$\pm 0.47$} & 69.47\tiny{$\pm 0.94$} \\ 
        & CODA-Prompt \cite{smith2023coda} & 3.84M & IN1K-Self & 75.22\tiny{$\pm 0.83$} & 84.17\tiny{$\pm 0.36$} & 60.77\tiny{$\pm 0.70$} & 68.86\tiny{$\pm 0.86$} \\ 
        \hdashline
        \multirow{4}*{\tabincell{c}{Ours}} 
        & \textcolor{gray}{SLCA~\cite{zhang2023slca}} & \textcolor{gray}{85.80M} & \textcolor{gray}{IN1K-Self} & \textcolor{gray}{85.27\tiny{$\pm 0.08$}} & \textcolor{gray}{89.51\tiny{$\pm 1.04$}} & \textcolor{gray}{68.07\tiny{$\pm 0.21$}} & \textcolor{gray}{73.04\tiny{$\pm 0.56$}} \\ 
        & \textcolor{gray}{SLCA++ (Full)} & \textcolor{gray}{85.80M} & \textcolor{gray}{IN1K-Self} & \textcolor{gray}{85.12\tiny{$\pm 0.15$}} & \textcolor{gray}{89.62\tiny{$\pm 1.11$}} & \textcolor{gray}{68.84\tiny{$\pm 0.18$}} & \textcolor{gray}{73.77\tiny{$\pm 0.55$}} \\
        & SL++ & 0.64M & IN1K-Self & 81.83\tiny{$\pm 0.29$} & 87.64\tiny{$\pm 1.09$} & 66.63\tiny{$\pm 0.95$} & 72.93\tiny{$\pm 0.96$}  \\
        & SLCA++ & 0.64M & IN1K-Self & \textbf{84.77}\tiny{$\pm 0.18$} & \textbf{89.53}\tiny{$\pm 0.98$} & \textbf{69.01}\tiny{$\pm 0.42$} & \textbf{74.75}\tiny{$\pm 0.69$} \\ 
        \hline
    \end{tabular}
    }
\label{table:cifar+inr}
\end{table*}

\begin{table*}[ht]
    \centering
    \caption{Experimental results for continual learning on Split CUB-200 and Split Cars-196. IN21K-Sup: supervised pre-training on ImageNet-21K. IN1K-Self: self-supervised pre-training on ImageNet-1K with MoCo v3 \cite{chen2021empirical}. All other continual learning methods based on Seq FT are reproduced according to their public codes with the proposed \textbf{Slow Learner} implemented. \textcolor{gray}{SLCA++ (Full)}: our full version that directly tuning all parameters in the backbone.}
    \footnotesize{
    \begin{tabular}{c|l|c|c||c|c||c|c}
    \hline
        & \multirow{2}{*}{Method} & \multirow{2}{*}{\#Params} & \multirow{2}{*}{Pre-trained} & \multicolumn{2}{c||}{Split CUB-200} & \multicolumn{2}{c}{Split Cars-196} \\ \cline{5-8}
        & & & & Last-Acc (\%) & Inc-Acc (\%)& Last-Acc (\%)& Inc-Acc (\%)\\ \hline
        Upper-Bound & Joint-Training & 85.80M & IN21K-Sup & 88.00\tiny{$\pm 0.34$} & -  & 80.31\tiny{$\pm 0.13$} & -  \\ 
        \hline
        \multirow{5}*{\tabincell{c}{Classical}} 
        & GDumb \cite{prabhu2020gdumb} & 85.80M & IN21K-Sup & 61.80\tiny{$\pm 0.77$} & 79.76\tiny{$\pm 0.18$} & 25.20\tiny{$\pm 0.84$} & 49.48\tiny{$\pm 0.74$} \\
        & DER++ \cite{buzzega2020dark}& 85.80M & IN21K-Sup & 77.42\tiny{$\pm 0.71$} & 87.61\tiny{$\pm 0.09$} & 60.41\tiny{$\pm 1.76$} & 75.04\tiny{$\pm 0.57$} \\  
        & BiC \cite{wu2019large}& 85.80M & IN21K-Sup & 81.91\tiny{$\pm 2.59$} & 89.29\tiny{$\pm 1.57$} & 63.10\tiny{$\pm 5.71$} & 73.75\tiny{$\pm 2.37$} \\ 
        & EWC \cite{kirkpatrick2017overcoming}& 85.80M  & IN21K-Sup & 68.32\tiny{$\pm 2.64$} & 79.95\tiny{$\pm 2.28$} & 52.50\tiny{$\pm 3.18$} & 64.01\tiny{$\pm 3.25$}  \\ 
        & LwF \cite{li2017learning}& 85.80M  & IN21K-Sup & 69.75\tiny{$\pm 1.37$} & 80.45\tiny{$\pm 2.08$} & 49.94\tiny{$\pm 3.24$} & 63.28\tiny{$\pm 1.11$}  \\
        \hdashline
        \multirow{4}*{\tabincell{c}{Param-Efficient}} 
        & L2P \cite{wang2022l2p}& 0.46M & IN21K-Sup & 62.21\tiny{$\pm 1.92$} & 73.83\tiny{$\pm 1.67$} & 38.18\tiny{$\pm 2.33$} & 51.79\tiny{$\pm 4.19$} \\ 
        & DualPrompt \cite{wang2022dualprompt}& 0.49M & IN21K-Sup & 66.00\tiny{$\pm 0.57$} & 77.92\tiny{$\pm 0.50$} & 40.14\tiny{$\pm 2.36$} & 56.74\tiny{$\pm 1.78$} \\ 
        & LAE (Adapter) \cite{gao2023unified} & 0.15M & IN21K-Sup & 77.48\tiny{$\pm 0.94$} & 85.83\tiny{$\pm 0.68$} & 52.47\tiny{$\pm 1.46$} & 64.08\tiny{$\pm 1.01$} \\ 
        & CODA-Prompt \cite{smith2023coda} & 3.84M & IN21K+IN1K-Sup & 72.63\tiny{$\pm 0.76$} & 80.54\tiny{$\pm 0.54$} & 44.89\tiny{$\pm 0.61$} & 58.91\tiny{$\pm 0.37$} \\

        \hdashline
        \multirow{4}*{\tabincell{c}{Ours}} 
        & \textcolor{gray}{SLCA~\cite{zhang2023slca}} & \textcolor{gray}{85.80M} & \textcolor{gray}{IN21K-Sup} & \textcolor{gray}{84.71\tiny{$\pm 0.40$}} & \textcolor{gray}{90.94\tiny{$\pm 0.68$}} & \textcolor{gray}{67.73\tiny{$\pm 0.85$}} & \textcolor{gray}{76.93\tiny{$\pm 1.21$}} \\
        & \textcolor{gray}{SLCA++ (Full)} & \textcolor{gray}{85.80M} & \textcolor{gray}{IN21K-Sup} & \textcolor{gray}{86.07\tiny{$\pm 0.06$}} & \textcolor{gray}{91.52\tiny{$\pm 0.67$}} & \textcolor{gray}{72.20\tiny{$\pm 0.44$}} & \textcolor{gray}{79.24\tiny{$\pm 0.58$}} \\
        & SL++ & 0.64M  & IN21K-Sup & 79.80\tiny{$\pm 0.67$} & 87.51\tiny{$\pm 0.74$} & 63.68\tiny{$\pm 0.29$} & 72.85\tiny{$\pm 0.66$} \\ 
        & SLCA++ & 0.64M & IN21K-Sup & \textbf{86.59}\tiny{$\pm 0.29$} & \textbf{91.63}\tiny{$\pm 0.72$} & \textbf{73.97}\tiny{$\pm 0.22$} & \textbf{79.46}\tiny{$\pm 0.80$} \\
        \hline \hline
        Upper-Bound & Joint-Training & 85.80M  & IN1K-Self & 79.55\tiny{$\pm 0.04$} & - & 74.52\tiny{$\pm 0.09$} & -  \\ 
        \hline
        \multirow{5}*{\tabincell{c}{Classical}} 
        & GDumb \cite{prabhu2020gdumb}& 85.80M  & IN1K-Self & 45.29\tiny{$\pm 0.97$} & 66.86\tiny{$\pm 0.63$} & 20.95\tiny{$\pm 0.42$} & 45.40\tiny{$\pm 0.66$} \\ 
        & DER++ \cite{buzzega2020dark}& 85.80M  & IN1K-Self  & 61.47\tiny{$\pm 0.32$} & 77.15\tiny{$\pm 0.61$} & 50.64\tiny{$\pm 0.70$} & 67.64\tiny{$\pm 0.45$} \\ 
        & BiC \cite{wu2019large}& 85.80M  & IN1K-Self & 74.39\tiny{$\pm 1.12$} & 82.13\tiny{$\pm 0.33$} & 65.57\tiny{$\pm 0.93$} & 73.95\tiny{$\pm 0.29$} \\ 
        & EWC \cite{kirkpatrick2017overcoming}& 85.80M & IN1K-Self & 61.36\tiny{$\pm 1.43$} & 72.84\tiny{$\pm 2.18$}  & 53.16\tiny{$\pm 1.45$} & 63.61\tiny{$\pm 1.06$} \\ 
        & LwF \cite{li2017learning}& 85.80M  & IN1K-Self & 61.66\tiny{$\pm 1.95$} & 73.90\tiny{$\pm 1.91$}  & 52.45\tiny{$\pm 0.48$} & 63.87\tiny{$\pm 0.31$}  \\ 
        \hdashline
        \multirow{4}*{\tabincell{c}{Param-Efficient}} 
        & L2P \cite{wang2022l2p}& 0.46M & IN1K-Self & 46.11\tiny{$\pm 1.09$}  & 67.27\tiny{$\pm 0.93$} & 36.29\tiny{$\pm 1.37$} & 50.47\tiny{$\pm 1.02$} \\ 
        & DualPrompt \cite{wang2022dualprompt}& 0.49M & IN1K-Self & 48.47\tiny{$\pm 1.31$} & 68.36\tiny{$\pm 1.22$} & 36.99\tiny{$\pm 1.64$} & 51.03\tiny{$\pm 1.63$} \\
        & LAE (Adapter) \cite{gao2023unified} & 0.15M & IN1K-Self & 53.72\tiny{$\pm 1.12$} & 69.92\tiny{$\pm 1.73$} & 41.16\tiny{$\pm 1.85$} & 56.97\tiny{$\pm 1.56$} \\ 
        & CODA-Prompt \cite{smith2023coda} & 3.84M & IN1K-Self & 50.22\tiny{$\pm 1.38$} & 68.77\tiny{$\pm 0.93$} & 38.81\tiny{$\pm 1.39$} & 53.86\tiny{$\pm 1.13$} \\ 
        \hdashline
        \multirow{4}*{\tabincell{c}{Ours}} 
        & \textcolor{gray}{SLCA~\cite{zhang2023slca}} & \textcolor{gray}{85.80M} & \textcolor{gray}{IN1K-Self} & \textcolor{gray}{73.01\tiny{$\pm 0.16$}} & \textcolor{gray}{82.13\tiny{$\pm 0.34$}} &  \textcolor{gray}{66.04\tiny{$\pm 0.08$}} & \textcolor{gray}{72.59\tiny{$\pm 0.04$}} \\
        & \textcolor{gray}{SLCA++ (Full)} & \textcolor{gray}{85.80M} & \textcolor{gray}{IN1K-Self} & \textcolor{gray}{74.35\tiny{$\pm 0.11$}} & \textcolor{gray}{81.95\tiny{$\pm 0.88$}} & \textcolor{gray}{68.04\tiny{$\pm 0.48$}} & \textcolor{gray}{74.02\tiny{$\pm 0.23$}} \\
        & SL++ & 0.64M  & IN1K-Self & 65.30\tiny{$\pm 1.15$} & 75.64\tiny{$\pm 1.81$} & 58.99\tiny{$\pm 0.91$} & 68.30\tiny{$\pm 0.28$} \\ 
        & SLCA++ & 0.64M  & IN1K-Self & \textbf{75.48}\tiny{$\pm 0.31$} & \textbf{82.94}\tiny{$\pm 0.73$} &  \textbf{69.71}\tiny{$\pm 0.10$} & \textbf{75.67}\tiny{$\pm 0.32$} \\ 
        \hline
    \end{tabular}
    }
\label{table:cub+cars}
\end{table*}

\section{Experiment}
\label{sec:exp}
In this section, we first briefly describe the experimental setups, and then present the experimental results. 

\subsection{Experimental Setups}

\textbf{Datasets}. 
Following L2P~\cite{wang2022l2p} and DualPrompt~\cite{wang2022dualprompt}, we adopt pre-training from ImageNet-21K dataset \cite{ridnik2021imagenet21k}, also known as the full ImageNet \cite{deng2009imagenet} consisting of 14,197,122 images with 21,841 classes. We also consider pre-training from ImageNet-1K dataset \cite{krizhevsky2012imagenet}, a subset of ImageNet-21K introduced for the ILSVRC2012 visual recognition challenge, consisting of 1000-class images.

To evaluate the performance of downstream continual learning, for class-incremental setting, we consider four representative benchmark datasets and randomly split each of them into 10 disjoint tasks:
The first two follow previous efforts \cite{wang2022l2p,wang2022dualprompt} and are relatively coarse-grained in terms of classification, while the last two are relatively fine-grained.
Specifically, CIFAR-100 dataset \cite{krizhevsky2009learning} consists of 100-class natural images with 500 training samples per class. ImageNet-R dataset \cite{hendrycks2021many} contains 200-class images, split into 24,000 and 6,000 images for training and testing (similar ratio for each class), respectively. Note that although the image categories of ImageNet-R are overlapped with ImageNet-21K, all images are out-of-distribution samples for the pre-train dataset, \textit{i.e.}, hard examples from ImageNet or newly collected data of different styles. It requires considerable adaptations of the pre-trained model, therefore serving as a challenging benchmark for continual learning. CUB-200 dataset \cite{wah2011caltech} includes 200-class bird images with around 60 images per class, 30 of which are used for training and the rest for testing. Cars-196 dataset \cite{krause20133d} includes 196 types of car images, split into 8,144 and 8,040 images for training and testing (similar ratio for each class), respectively.
We also consider the domain-incremental setting following S-Prompts~\cite{wang2022s} and evaluate our approach on DomainNet~\cite{peng2019moment}, a dataset with 345 categories from 6 different domains, counting a total of more than 600,000 images.

\textbf{Evaluation Metrics}. We present the average accuracy of all classes after learning the last task, denoted as Last-Acc (equivalent to ``Avg. Acc'' in \cite{wang2022l2p,wang2022dualprompt}). We also compute the average accuracy of the classes ever seen after learning each incremental task and then present their average after learning the last task, denoted as Inc-Acc.

\textbf{Implementations}.
Following previous efforts \cite{wang2022l2p,wang2022dualprompt}, we adopt a pre-trained ViT-B/16 backbone for all baselines. 
For recent works on CLPT, such as L2P \cite{wang2022l2p}, DualPrompt \cite{wang2022dualprompt}, CODA-Prompt \cite{smith2023coda}, etc., we follow their official implementation and employ the Adam optimizer for training. For our approach and other continual learning methods based on Seq FT, an SGD optimizer is used, with the same batch size of 128. Our SL adopts a learning rate of 0.0001 for the representation layer (0.001 for our hybrid version) and 0.01 for the classification layer.

\textbf{Baselines}.
We adopt joint training as the upper bound performance and consider continual learning baselines with or without replaying old training samples. 
As for the former, a memory buffer of 1000 images is maintained, and we evaluate three representative replay-based methods such as BiC \cite{wu2019large}, GDumb \cite{prabhu2020gdumb} and DER++ \cite{buzzega2020dark}. As for the latter, we evaluate representative regularization-based methods such as EWC \cite{kirkpatrick2017overcoming} and LwF \cite{li2017learning}, and prompt-based methods such as L2P \cite{wang2022l2p}, DualPrompt \cite{wang2022dualprompt} and CODA-Prompt~\cite{smith2023coda}. We also evaluate LAE~\cite{gao2023unified}, a concurrent parameter-efficient fine-tuning method implemented with adapter.
Note that Seq FT usually serves as the lower bound performance of continual learning, but we observe that simply adjusting the learning rate (\textit{i.e.}, using the proposed SL) makes it a surprisingly strong baseline for CLPT.

\begin{table}[h]
    \small
    \caption{Domain-incremental learning on Split DomainNet. We follow the same implementation as S-iPrompts~\cite{wang2022s} for all experiments. ImageNet-21K pre-training is used for all methods.}
    \vspace{-0.1cm}
    \label{table:domainet}
    \begin{center}
    \begin{tabular}{l|c|c}
    \hline 
     Method  & \#Params & Last-Acc (\%)   \\
    \hline
     EWC~\cite{kirkpatrick2017overcoming}  & 85.80M & 47.62  \\
     LwF~\cite{li2017learning} & 85.80M & 49.19  \\
     CaSSLe (Supervised) \cite{fini2022self} & 85.80M & 55.90 \\
     L2P \cite{wang2022l2p} & 0.46M & 40.15  \\
     S-iPrompts \cite{wang2022s} & 0.05M & 50.62  \\
     \hline
     SLCA++ & 0.64M & \textbf{59.45} \\
     \textcolor{gray}{SLCA++ (Full)} & \textcolor{gray}{85.80M} & \textcolor{gray}{61.18} \\
    \hline
    \end{tabular}
    \end{center}
    \vspace{-6mm}
    \end{table}

\subsection{Overall Performance}
All continual learning methods based on Seq FT (\textit{i.e.}, the classical baselines) in Table~\ref{table:cifar+inr} and~\ref{table:cub+cars} are equipped with our \textbf{Slow Learner} (SL) for fair comparison.
For continual learning on relatively coarse-grained classification benchmarks, such as Split CIFAR-100 and Split ImageNet-R in Table~\ref{table:cifar+inr} (also shown in Fig.~\ref{img21k_sup_all}), the proposed SL can substantially enhance the final performance. With the help of Classifier Alignment (CA) and its Logit Normalization (LN), SLCA++ clearly outperforms state-of-the-art baselines~\cite{wang2022l2p,wang2022dualprompt,smith2023coda,gao2023unified}, and can almost reach the joint training performance (the performance gap is less than \textbf{2\%} under supervised pre-training and \textbf{4\%} under self-supervised pre-training). 
For continual learning on relatively fine-grained classification benchmarks, such as Split CUB-200 and Split Cars-196 in Table~\ref{table:cub+cars}, SLCA++ consistently surpasses all other baselines by a large margin, which validates its generality. 
In Table~\ref{table:domainet}, we also evaluate SLCA++ on Split DomainNet benchmark~\cite{peng2019moment} for domain-incremental learning. 
Our SLCA++ with either full parameter tuning or parameter-efficient fine-tuning achieves remarkably better performance than all other baselines.

Regarding the results of self-supervised pre-training (\textit{i.e.}, MoCov3 in our experiments), prompt-based methods perform far below the joint training. This is not surprisingly as the visual prompt tuning~\cite{jia2022visual} has been shown to perform much worse than full parameter tuning when adapting self-supervised pre-trained models to downstream tasks. LAE~\cite{gao2023unified} with adapter show improvement over the prompt-based method, but still underperform SLCA++, especially on fine-grained datasets. Under the same parameter-efficient paradigm, our SLCA++ far exceeds other methods, while showing comparable results on Split CIFAR-100 and Split ImageNet-R compared to SLCA++ with full parameter tuning. More interestingly, a better performance is achieved with our hybrid version on fine-grained datasets like Split CUB-200 and Split Cars-196.

Besides, it is worth noting that different replay-based methods (w/ SL) behave differently in CLPT. 
In general, BiC \cite{wu2019large}, which adds a bias correction layer to mitigate imbalance between old and new classes after the learning of representation layer using both newly added and old training samples, receives the most substantial improvements. 
While GDumb \cite{prabhu2020gdumb}, which simply uses limited training samples storing in the memory to train a newly initialized model at test time, has difficulty in optimization and performs the worst (especially on fine-grained datasets). Therefore, CLPT brings advantages but also challenges for the area of continual learning.

\begin{table*}[ht]
	\centering
    \caption{Ablation study of major components in SLCA++ (Full). Here we present the Last-Acc (\%) after continual learning of all tasks.
    $^{\dag}$The reproduced performance of Seq FT is slightly different from the reported one in~\cite{wang2022l2p,wang2022dualprompt} due to the use of different random seeds. SL: Slow Learner; LN: Logit Normalization; CA: an implementation of Classifier Alignment without LN. SCE: Symmetric Cross-Entropy.} 

    \footnotesize{
 { 
	\begin{tabular}{c|l|c|c|c|c|c}
	 \hline
       & Method & Pre-trained & Split CIFAR-100 & Split ImageNet-R & Split CUB-200 & Split Cars-196 \\
        \hline
       \multirow{2}*{\tabincell{c}{Baseline}}
       &Seq FT $^{\dag}$&IN21K-Sup & 41.77\tiny{$\pm 13.8$} & 26.95\tiny{$\pm 11.8$} & 40.02\tiny{$\pm 1.08$} & 27.57\tiny{$\pm 1.79$} \\ 
       &w/ Fixed $\theta_{rps}$ &IN21K-Sup & 63.75\tiny{$\pm 0.67$}  & 34.64\tiny{$\pm 14.3$} & 60.44\tiny{$\pm 1.80$} & 24.51\tiny{$\pm 6.90$} \\
       \hline
      \multirow{5}*{\tabincell{c}{Ours}}
       &w/ SL &IN21K-Sup &88.86\tiny{$\pm 0.83$} &71.80\tiny{$\pm 1.45$} &68.07\tiny{$\pm 1.09$} &49.74\tiny{$\pm 1.25$} \\     
       &w/ Fixed $\theta_{rps}$+CA & IN21K-Sup & 75.64\tiny{$\pm 0.26$} & 50.73\tiny{$\pm 0.21$} & 82.71\tiny{$\pm 0.14$} & 54.45\tiny{$\pm 0.16$}   \\ 
       &w/ Fixed $\theta_{rps}$+CA+LN & IN21K-Sup & 75.62\tiny{$\pm 0.21$} & 51.83\tiny{$\pm 0.34$} & 83.65\tiny{$\pm 0.18$} & 53.43\tiny{$\pm 0.09$} \\   
       &w/ SL+CA &IN21K-Sup & 90.70\tiny{$\pm 0.52$} & 74.41\tiny{$\pm 0.51$} & 83.20\tiny{$\pm 0.19$} & 67.90\tiny{$\pm 0.53$}  \\
       &w/ SL+CA+LN &IN21K-Sup & 91.53\tiny{$\pm 0.28$}  &77.00\tiny{$\pm 0.33$}  & 84.71\tiny{$\pm 0.40$}  & 67.73\tiny{$\pm 0.85$} \\
       &w/ SL+SCE+CA+LN & IN21K-Sup &\textbf{91.69}\tiny{$\pm 0.15$}  & \textbf{79.78}\tiny{$\pm 0.16$}  & \textbf{86.07}\tiny{$\pm 0.06$}  & \textbf{72.20}\tiny{$\pm 0.44$} \\ 
       \hline
       \hline
        \multirow{2}*{\tabincell{c}{Baseline}}  
       &Seq FT &IN1K-Self & 27.99\tiny{$\pm 5.16$} & 45.84\tiny{$\pm 4.19$} & 45.35\tiny{$\pm 1.38$} &  35.96\tiny{$\pm 2.04$} \\ 
       &w/ Fixed $\theta_{rps}$ &IN1K-Self & 77.30\tiny{$\pm 0.56$} & 51.97\tiny{$\pm 0.17$} & 55.54\tiny{$\pm 1.55$} & 43.16\tiny{$\pm 0.12$} \\ 
       \hline
      \multirow{5}*{\tabincell{c}{Ours}}
       &w/ SL &IN1K-Self &81.47\tiny{$\pm 0.55$} &64.43\tiny{$\pm 0.44$} &61.67\tiny{$\pm 1.37$} &52.91\tiny{$\pm 1.53$} \\  
       &w/ Fixed $\theta_{rps}$+CA  & IN1K-Self & 81.83\tiny{$\pm 0.12$} & 55.59\tiny{$\pm 0.21$} & 70.67\tiny{$\pm 0.02$} & 57.01\tiny{$\pm 0.07$} \\ 
       &w/ Fixed $\theta_{rps}$+CA+LN & IN1K-Self & 81.95\tiny{$\pm 0.17$} & 56.47\tiny{$\pm 0.23$} & 72.97\tiny{$\pm 0.17$} & 63.00\tiny{$\pm 0.21$} \\       
       &w/ SL+CA &IN1K-Self & 84.64\tiny{$\pm 0.21$} & 67.54\tiny{$\pm 0.29$} & 72.52\tiny{$\pm 0.06$} & 64.80\tiny{$\pm 0.20$} \\    
       &w/ SL+CA+LN &IN1K-Self &\textbf{85.27}\tiny{$\pm 0.08$} & 68.07\tiny{$\pm 0.21$}  & 73.01\tiny{$\pm 0.16$}  & 66.04\tiny{$\pm 0.08$} \\
       &w/ SL+SCE+CA+LN &IN1K-Self &85.12\tiny{$\pm 0.15$}  &\textbf{68.84}\tiny{$\pm 0.18$}  & \textbf{74.35}\tiny{$\pm 0.11$}  & \textbf{68.04}\tiny{$\pm 0.48$} \\
       \hline
	\end{tabular}
	}
}
	\label{table:ablation}
\end{table*}

\begin{table*}[ht]
    \vspace{-3mm}
    \centering
    \caption{Ablation study of varying tuning parameters. All entries tune the backbone in a Seq FT manner. X-only means only optimizing parameter within all X components. Partial-1 means only tuning the last transformer block. LoRA-x means using LoRA adaptation with rank x. Hybrid denotes our hybrid slow learner. We chose \colorbox{black!15}{Hybrid (LoRA-4)} as our default choice when comparing with other works.} 

    \footnotesize{
 { 
	\begin{tabular}{c|l|c|c|c|c|c}
	 \hline

        Tuning Type & Method & \#Params & Split CIFAR-100 & Split ImageNet-R & Split CUB-200 & Split Cars-196 \\

        \hline
       \multirow{3}*{\tabincell{c}{Classical}}  
        & Full & 85.80M & \textbf{91.69}\tiny{$\pm 0.15$}  & \textbf{79.78}\tiny{$\pm 0.16$}  & 86.07\tiny{$\pm 0.06$}  & 72.20\tiny{$\pm 0.44$} \\ 
        & Mlp-only & 56.67M & 91.49\tiny{$\pm 0.11$}  & 79.26\tiny{$\pm 0.17$} & 84.88\tiny{$\pm 0.09$} & 69.65\tiny{$\pm 0.53$} \\
        & Attn-only & 21.26M & 91.61\tiny{$\pm 0.09$} & 77.88\tiny{$\pm 0.21$} & 84.66\tiny{$\pm 0.16$} & 70.21\tiny{$\pm 0.49$} \\
        \hdashline
        \multirow{7}*{\tabincell{c}{Param-Efficient}}  
        & Partial-1 & 7.09M & 75.99\tiny{$\pm 0.21$} & 54.59\tiny{$\pm 0.18$} & 84.28\tiny{$\pm 0.13$} & 51.31\tiny{$\pm 0.39$} \\
        & Norm-only & 0.04M & 89.47\tiny{$\pm 0.16$} & 73.10\tiny{$\pm 0.10$} & 84.98\tiny{$\pm 0.15$} & 64.97\tiny{$\pm 0.51$} \\
        & bias-only & 0.08M & 91.12\tiny{$\pm 0.11$} & 73.95\tiny{$\pm 0.23$} & 84.81\tiny{$\pm 0.28$} & 66.84\tiny{$\pm 0.44$} \\
        & LoRA-only (LoRA-1) & 0.13M & 91.48\tiny{$\pm 0.15$} & 76.30\tiny{$\pm 0.26$} & 83.59\tiny{$\pm 0.22$} & 65.91\tiny{$\pm 0.50$} \\       
        & Hybrid (LoRA-1) & 0.25M & 91.61\tiny{$\pm 0.19$} & 76.69\tiny{$\pm 0.19$} & 85.39\tiny{$\pm 0.19$} & 69.41\tiny{$\pm 0.37$} \\
        \rowcolor{black!15}
        & Hybrid (LoRA-4) & 0.64M & 91.46\tiny{$\pm 0.18$} & 78.09\tiny{$\pm 0.22$}  & \textbf{86.59}\tiny{$\pm 0.29$} & \textbf{73.97}\tiny{$\pm 0.22$} \\
        & \textcolor{gray}{Hybrid (LoRA-4 w/o init.)} & \textcolor{gray}{0.64M} & \textcolor{gray}{90.16\tiny{$\pm 0.38$}} & \textcolor{gray}{77.47\tiny{$\pm 0.14$}} & \textcolor{gray}{85.19\tiny{$\pm 0.34$}} & \textcolor{gray}{65.37\tiny{$\pm 0.27$}} \\

       \hline
	\end{tabular}
	}
}
	\label{table:ablation_peft}
\end{table*}

\subsection{Ablation Study}
\textbf{Effect of Slow Learner (SL)}. 
We start from two basic baselines, including sequential fine-tuning with a uniform learning rate of 0.005 for both representation and classification layers, and continually learning classifier with a fixed representation layer, denoted as \textit{Seq FT} and \textit{(Seq FT) w/ Fixed $\theta_{rps}$} in Table~\ref{table:ablation}, respectively.
Although \textit{(Seq FT) w/ Fixed $\theta_{rps}$} performs better than \textit{Seq FT}, it is significantly inferior to \textit{Seq FT w/ SL}, indicating the \emph{necessity} of updating the representation layer while using a properly reduced learning rate to mitigate the progressive overfitting problem. 
Besides, we can see that while the benefits of SL are significant, the improvement is slightly limited in fine-grained datasets. Specifically, for ImageNet-21K supervised pre-training, the improvements of SL are \textbf{47.09\%}, \textbf{44.85\%} on Split~CIFAR-100 and Split~ImageNet-R but \textbf{28.05\%} and \textbf{22.17\%} for Split~CUB-200 and Split~Cars-196. As indicated by Fig.~\ref{img21k-cifar100-probe}, a classifier alignment strategy is required. 

\textbf{Effect of Classifier Alignment (CA)}.
We first apply CA (as well as LN) on the \textit{Seq FT w/ Fixed $\theta_{rps}$} baseline. Surprisingly, it brings substantial improvements especially on fine-grained datasets (60.44\% and 28.92\% on Split~CUB-200 and Split~Cars-196 with IN21K-Sup pre-training), demonstrating it strong capacity for solving the above mis-alignment problem. Moreover, CA+LN also works well with our SL, compensating improvement loss on fine-grained datasets. In details, it brings \textbf{2.67\%}, \textbf{5.20\%}, \textbf{19.64\%} and \textbf{17.99\%} on Split~CIFAR-100, Split~ImageNet-R, Split~CUB-200 and Split~Cars-196, respectively over the \textit{Seq FT} baseline with IN21K-Sup pre-training. With IN1K-Self pre-training, our CA and CA+LN exhibit a similar trend of improvement. 

Note that our CA is operated in a \emph{post-hoc} fashion rather than aligning the classifier during representation learning~\cite{zhu2021pass,hayes2020remind,hayes2020lifelong}, as the latter would \emph{worsen} the performance (\eg, by 27.89\% on Split~CIFAR-100) in our experiments.

\textbf{Effect of Symmetric Cross-Entropy}. 
With the Symmetric Cross-Entropy (SCE) objective, we can further improve the performance of our approach, especially on Split Cars-196, which is considered the most challenging benchmark in all experiments and has the largest domain shift compared to the pre-training dataset. The results are consistent with our analysis in Sec.~\ref{sec:sce}, where SCE balances the gradients of high and low confidence samples to improve stability when training on datasets with more hard samples.

\textbf{Effect of Hybrid Slow Learner}. Table~\ref{table:ablation_peft} illustrates the transformation of our approach into a parameter-efficient version. We start with tuning different parts of the backbone to transfer pre-trained knowledge for downstream continual learning. Tuning different parts has varying effects on different datasets, demonstrating that there is \emph{no one-size-fits-all solution} when only tuning a part of the backbone. Direcly adopting advanced parameter-efficient fine-tuning techniques like LoRA~\cite{hu2021LoRA} outperforms other parameter-efficient parts on Split ImageNet-R but falls behind them on fine-grained tasks. By combining them together as Hybrid Slow Learner, we achieve comparable or even better performance on fine-grained datasets compared to the full parameter tuning counterpart, providing a robust and strong PEFT-based solution for CLPT through Seq FT without the need of parameter selection or incremental parameter addition.

\subsection{Discussion}
\textbf{Pre-training Paradigm.}
Constructing strong pre-trained models typically requires large amounts of pre-training data, while the extensive annotation is scarce and expensive, making self-supervised pre-training a more realistic option than supervised pre-training. However, most of the existing CLPT studies only consider the use of strong supervised pre-training. 
Our initial investigation into \textit{continual learning with self-supervised pre-training} indicates that state-of-the-art methods for CLPT tend to face severe challenges with it. Given its practical significance and inherent difficulty, we suggest future CLPT studies to direct more effort into this avenue. On the other hand, we also suggest subsequent work to develop self-supervised pre-training methods that are better suited for downstream continual learning, and potentially use this as a criterion to evaluate the progress of self-supervised learning.

\textbf{Scalability}. We further discuss the scalability of our approach.
First, the generated features are only used to align the output layer at test time rather than training the entire backbone, thus the \emph{computation} is efficient and not accumulated in continual learning (e.g., ranging from only 0.67\% to 5\% of the total running time for various benchmarks).
Second, the class-wise covariance $\Sigma_c$ can be further simplified as by an global $\Sigma$ for all classes by momentum updating $\Sigma=\gamma\Sigma_{c-1}+(1-\gamma)\Sigma_{c}$ where $\gamma$ is the momentum value (\textit{e.g.,} 0.9), or the variance $\sigma_c^2 \in \mathbb{R}^{d}$. When replacing $\Sigma_c$ with $\sigma_c^2$, it only results in a tolerable performance degradation (e.g., up to only 0.61\% on Split CIFAR-100 and 0.83\% on Split ImageNet-R), while the \emph{storage} of $\mu_c$ and $\sigma_c^2$ for 100 classes corresponds to only 0.18\% parameters of the ViT-B/16 backbone, which is clearly lightweight. 

\textbf{Learning Rate for CL.} 
One of our most interesting findings in this work is that, using a properly adapted learning rate has been sufficiently effective for representation learning on a continual basis, which can contribute to many relevant topics.
For example, as pre-training data is often massive and incrementally collected, the problem of continual pre-training has received growing interest, especially for large language models (LLMs)~\cite{achiam2023gpt,anil2023palm}. We note that a majority of recent efforts on this topic proposed to selectively adjusting the learning rate for sequential training of all parameters instead of using specialized regularization strategies~\cite{guptacontinual,winata2023overcoming,ibrahim2024simple}, consistent with our results. 
This is possibly because the distribution of pre-training data is extremely complex, which limits the application of continual learning methods that are specialy designed. 
In contrast, adjusting the learning rate relies on clearly simplified prior assumptions, making it work better for continual learning in the context of pre-training. 
To the best of our knowledge, SLCA / SLCA++ is one of the first studies to explicitly analyze the impact of learning rate on CLPT, which may facilitate a series of subsequent work to develop more adaptive CLPT methods based on our approach.

\section{Conclusion}
In this work, we present the advanced Slow Learner with Classifier Alignment (SLCA++), a simple but effective approach for the problem of continual learning with pre-training (CLPT). This approach stems from our in-depth analysis of the progressive overfitting issue in CLPT, where using a uniformly large learning rate to update the entire model tends to compromise the benefits of both stability and generalizability from pre-training. In response, the Slow Learner (SL) demonstrates that using a selectively reduced learning rate for sequential fine-tuning can almost address this challenging issue in the representation layer, regardless of different pre-training paradigms and downstream granularity. Also, the Classifier Alignment (CA) helps to resolve the problem of sub-optimal classification layer by employing feature statistics. 
The combination of SL and CA unleash the hidden power of sequential fine-tuning for CLPT. We have further improved the efficacy and efficiency of our approach with Symmetric Cross-Entropy (SCE) and a parameter-efficient version called Hybrid Slow Learner (Hybrid SL), respectively. With all these designs, our approach achieves state-of-the-art results across a variety of CLPT benchmarks. 
Possible future work includes exploring better self-supervised pre-training for downstream continual learning as well as exploring SLCA++ in more scenarios, such as continual learning of multi-modal and embodied tasks.


{\small
\bibliographystyle{ieee_fullname}
\bibliography{egpaper_final.bib}
}

\vspace{-35pt}
\begin{IEEEbiography}
[{\includegraphics[width=1in,height=1.25in,clip,keepaspectratio]{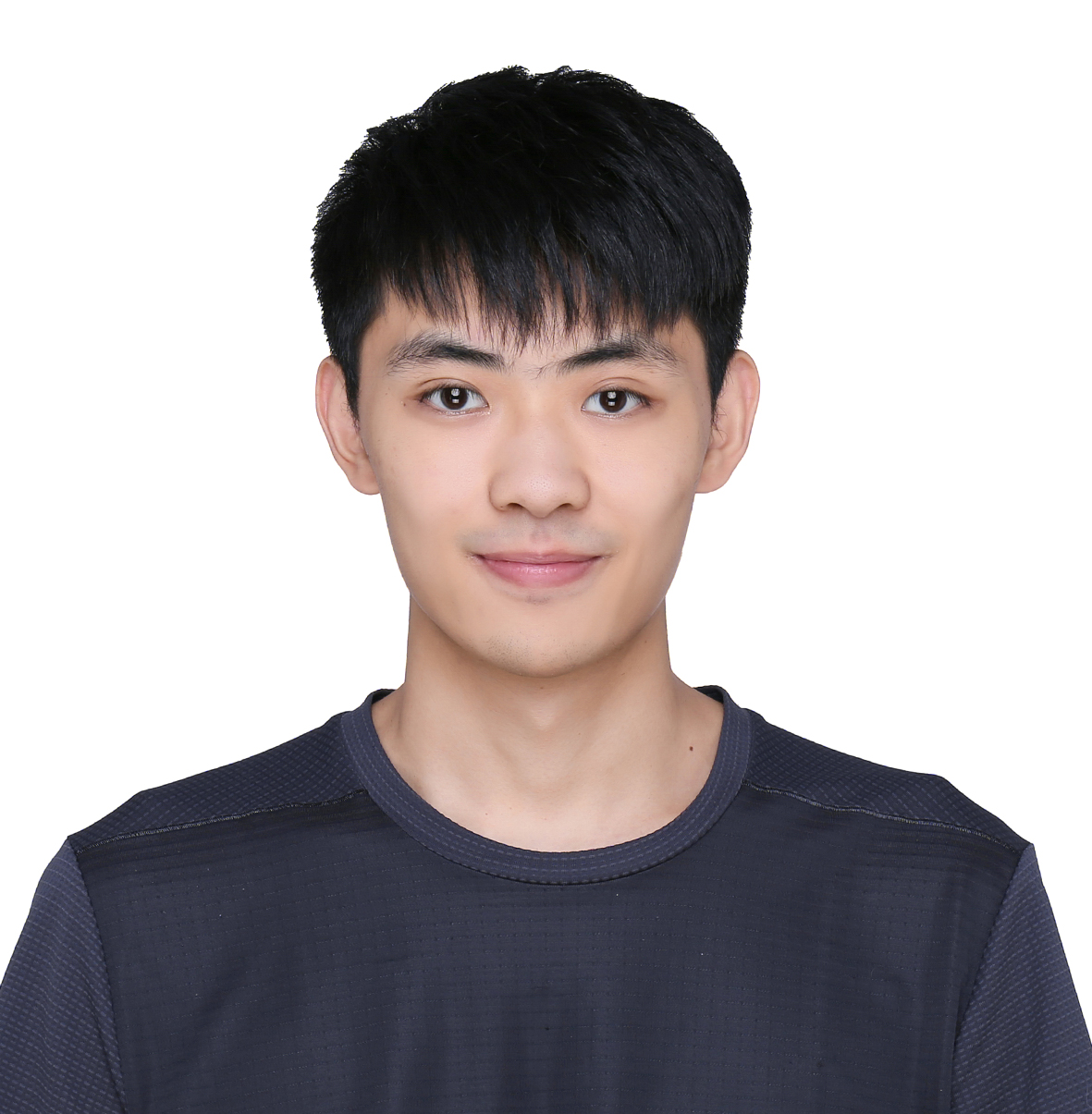}}]{Gengwei Zhang} is currently working toward the Ph.D. degree at the Faculty of Engineering and Information Technology, University of Technology Sydney, Australia, supervised by Prof. Ling Chen and Prof. Yunchao Wei. He received his Bachelor's degree from Sun Yat-Sen University, China, in 2019. His research interests focus on enhancing the learning capabilities of computer vision models using machine learning techniques such as continual learning, few-shot learning, and self-supervised learning.
\end{IEEEbiography}
\vspace{-35pt}
\begin{IEEEbiography}
[{\includegraphics[width=1in,height=1.25in,clip,keepaspectratio]{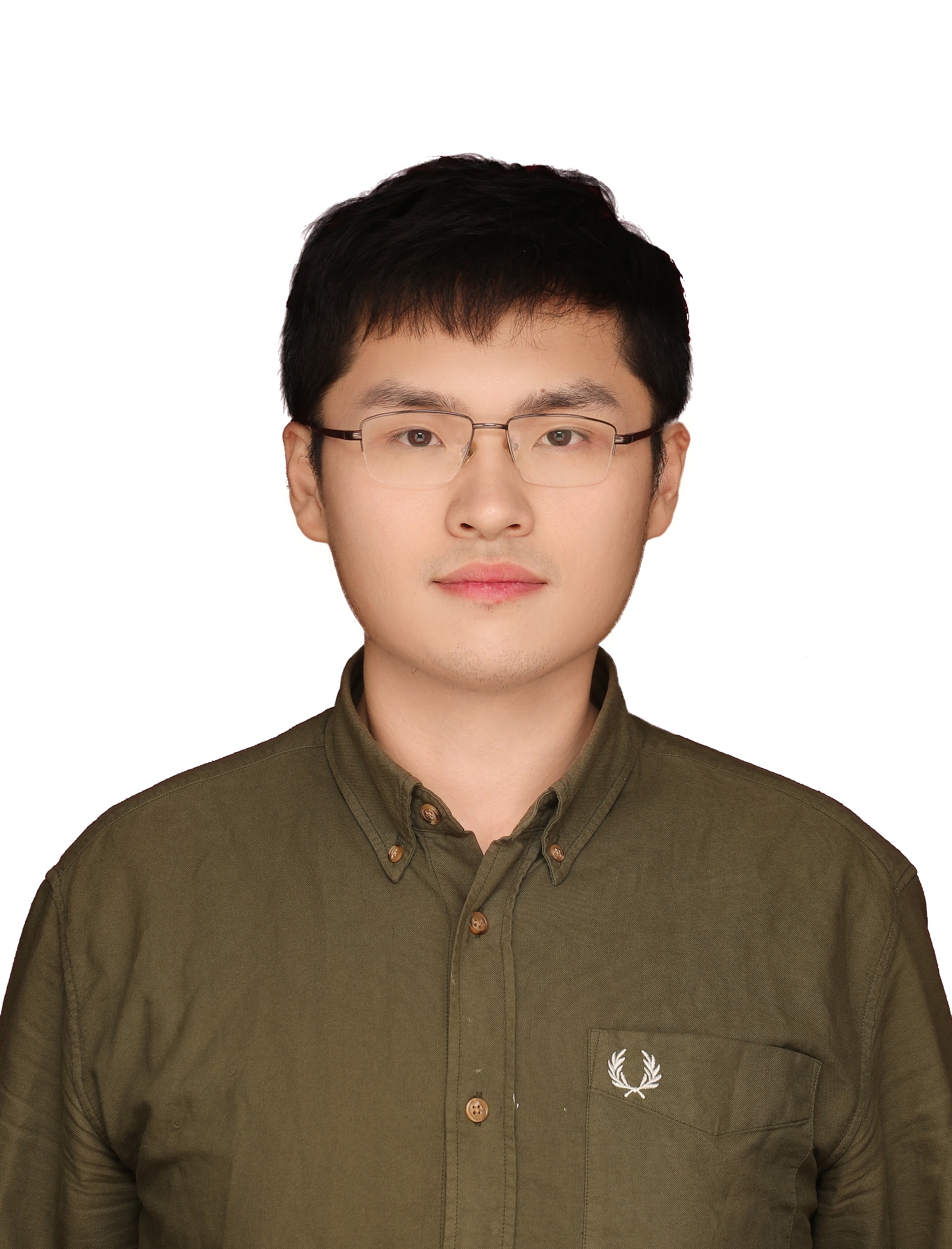}}]{Liyuan Wang}
is currently a postdoctoral researcher in Tsinghua University, working with Prof.~Jun Zhu at the Department of Computer Science and Technology. Before that, he received the B.S. and Ph.D. degrees from Tsinghua University. His research interests include continual learning, incremental learning, lifelong learning and brain-inspired AI. His work in continual learning has been published in major conferences and journals in related fields, such as Nature Machine Intelligence, IEEE TPAMI, IEEE TNNLS, NeurIPS, ICLR, CVPR, ICCV, ECCV, etc.
\end{IEEEbiography}
\vspace{-35pt}
\begin{IEEEbiography}
[{\includegraphics[width=1in,height=1.25in,clip,keepaspectratio]{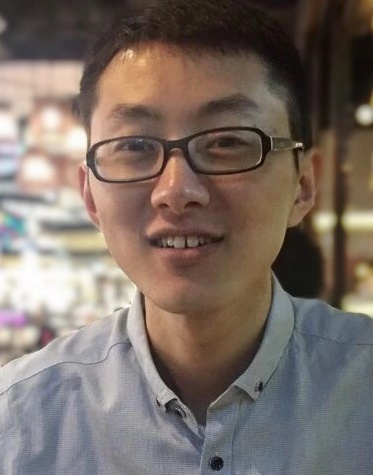}}]{Guoliang Kang}
is currently a Professor at Beihang University. He received his PhD degree from the University of Technology Sydney in 2019. He used to be a Postdoctoral Research Associate at Carnegie Mellon University and the University of Texas, Austin. His research interests include deep learning, computer vision, transfer learning, etc.
\end{IEEEbiography}
\vspace{-35pt}
\begin{IEEEbiography}
[{\includegraphics[width=1in,height=1.25in,clip,keepaspectratio]{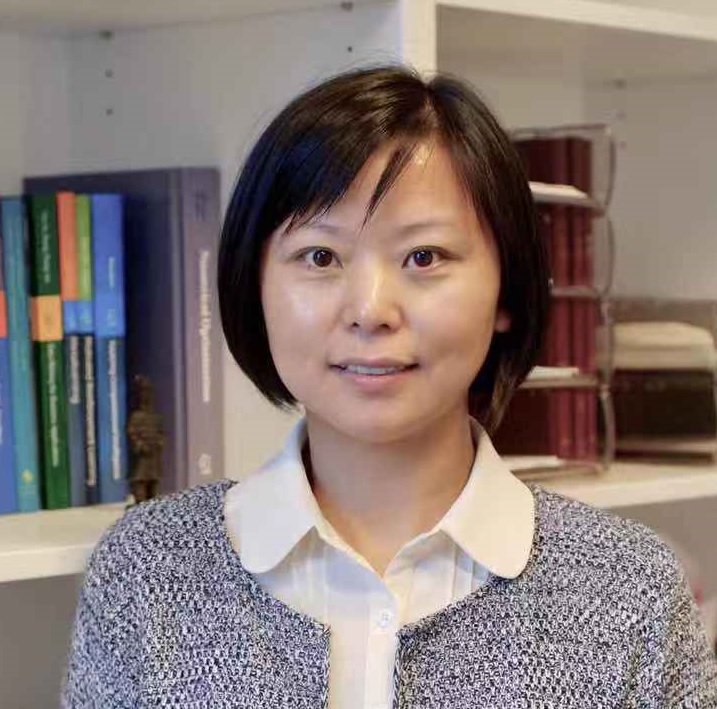}}]{Ling Chen}
is a Professor with the Australian Artificial Intelligence Institute (AAII), University of Technology Sydney. She received PhD from Nanyang Technological University, Singapore. Her research area is machine learning and data mining. Her recent research focuses on  developing moral aligned agents for text-based games, improving the robustness of dialogue systems, and anomaly detection from graph structured data. Her papers appear in major journals and conferences including IEEE TPAMI, IEEE TNNLS, NeurIPS and ICLR. She is an editorial member of the Elsevier Journal of Data and Knowledge Engineering, the Springer Journal of Data Science and Analytics, and the IEEE Journal of Social Computing. 
\end{IEEEbiography}
\vspace{-35pt}
\begin{IEEEbiography}
[{\includegraphics[width=1in,height=1.25in,clip,keepaspectratio]{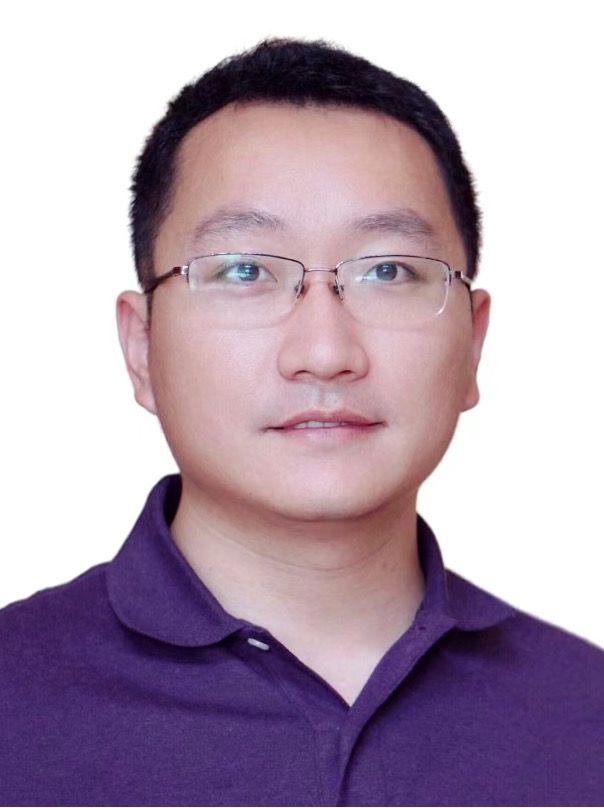}}]{Yunchao Wei}
is currently a full professor at Beijing Jiaotong University. Previously, he held positions at the National University of Singapore, the University of Illinois at Urbana-Champaign, and the University of Technology Sydney. He has published over 100 papers in top-tier conferences and journals, with more than 22,000 Google Scholar citations. His research interests include visual recognition with imperfect data, multi-modal perception, and generative AI.
\end{IEEEbiography}

\clearpage

\begin{appendices}

\begin{table*}[ht]
\hsize=\textwidth
   \centering
	\smallskip
 \caption{Continual learning performance with different learning rates of the representation layer. Here we present the Last-Acc (\%) after continual learning of all classes. IN21K-Sup: supervised pre-training on ImageNet-21K. IN1K-Self: self-supervised pre-training on ImageNet-1K with MoCo v3 \cite{chen2021empirical}. The column labeled by $^{\dag}$ uses the same learning rate of 0.005 for the entire model, while the others use a learning rate of 0.01 for the classification layer.} 
 { 
	\begin{tabular}{c|c|c|c|c|c|c|c}
	 \hline
       Benchmark & Pre-trained & 0.005$^{\dag}$ & 0.001 & 0.0001 & 0.00001 & 0.000001 & Fixed $\theta_{rps}$\\
        \hline
        Split CIFAR-100 &IN21K-Sup &44.77 \small{$\pm 13.8$} &83.04\small{$\pm 1.46$} &88.86\small{$\pm 0.83$} &88.81\small{$\pm 0.46$} &85.11\small{$\pm 0.42$} &63.75\small{$\pm 0.67$}\\ 
        Split ImageNet-R &IN21K-Sup &26.95 \small{$\pm 11.8$} &70.38\small{$\pm 0.80$} &71.80\small{$\pm 1.45$} &62.64\small{$\pm 2.35$} &53.57\small{$\pm 4.33$} &34.64\small{$\pm 14.3$}\\  
        Split CUB-200 &IN21K-Sup &40.02 \small{$\pm 1.08$} &60.02\small{$\pm 1.24$} &68.07\small{$\pm 1.09$} &66.58\small{$\pm 3.93$} &64.38\small{$\pm 3.36$} &60.44\small{$\pm 1.80$}\\  
        Split Cars-196 &IN21K-Sup &27.57 \small{$\pm 1.79$} &15.74\small{$\pm 26.3$} &49.74\small{$\pm 1.25$} &30.66\small{$\pm 9.01$} &24.85\small{$\pm 7.90$} &24.51\small{$\pm 6.90$}\\  
        \hline
        \hline
        Split CIFAR-100 &IN1K-Self &27.99 \small{$\pm 5.16$} &81.49\small{$\pm 0.75$} &81.47\small{$\pm 0.55$} &81.57\small{$\pm 0.14$} &78.61\small{$\pm 0.29$} &77.30\small{$\pm 0.56$}\\ 
        Split ImageNet-R &IN1K-Self &45.84 \small{$\pm 4.19$} &68.72\small{$\pm 0.48$} &64.43\small{$\pm 0.44$} &59.19\small{$\pm 0.33$} &54.54\small{$\pm 0.32$} &51.97\small{$\pm 0.17$}\\  
        Split CUB-200 &IN1K-Self &45.35 \small{$\pm 1.38$} &68.58\small{$\pm 1.16$} &61.67\small{$\pm 1.37$} &56.46\small{$\pm 1.86$} &55.10\small{$\pm 2.13$} &55.54\small{$\pm 1.55$}\\  
        Split Cars-196 &IN1K-Self &35.96 \small{$\pm 2.04$} &58.39\small{$\pm 2.31$} &52.91\small{$\pm 1.61$} &43.64\small{$\pm 0.73$} &41.74\small{$\pm 0.23$} &43.16\small{$\pm 0.12$}\\  
        \hline
	\end{tabular}
	}
	\label{table:learning_rate}
      \vspace{+0.1cm}
\end{table*}

\begin{figure}[h]
    \centering
    \includegraphics[width=0.95\linewidth]{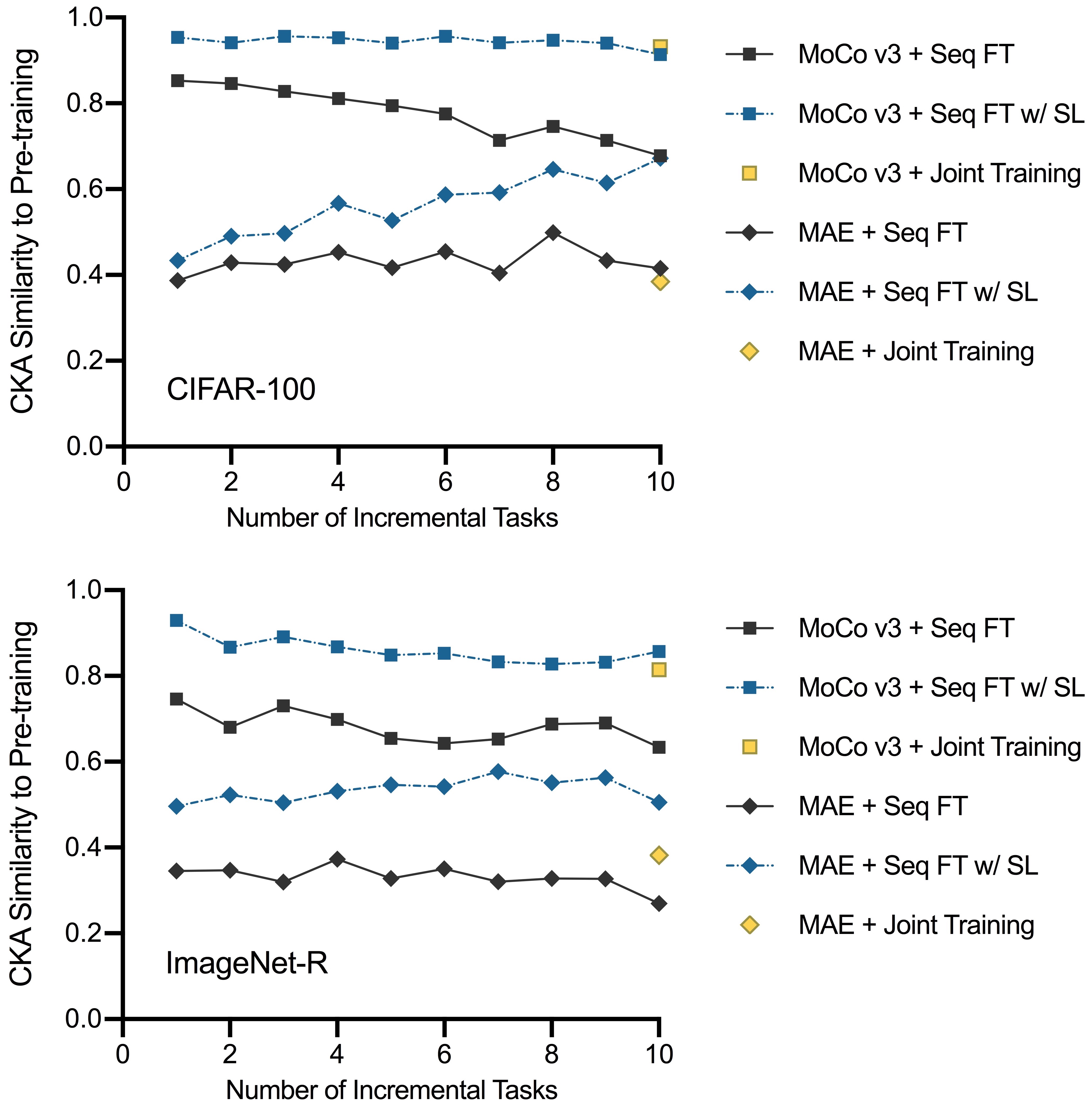}
    \caption{CKA similarity of pre-trained representations before and after learning downstream tasks.
    }
    \label{ssl_cka_cl_pt}
\end{figure}

\section{Implementation Details}

For class-incremental setting, all baselines follow an implementation similar to the one described in \cite{wang2022l2p,wang2022dualprompt}. Specifically, a pre-trained ViT-B/16 backbone is adopt for all methods. Adam optimizer is used for prompting-based methods~\cite{wang2022l2p,wang2022dualprompt,smith2023coda} as well as for LAE~\cite{gao2023unified}. An SGD optimizer is utilized for other baselines and ours, with the same batch size of 128. The original implementation of \cite{wang2022l2p,wang2022dualprompt} adopts a constant learning rate of 0.005 for all baselines, while our slow learner using 0.0001 for our full version and 0.001 for hybrid version to update representation layer, and 0.01 for the classification layer. In practice, we observe that supervised pre-training usually converges faster than self-supervised pre-training in downstream continual learning. Therefore, for supervised pre-training, we train all baselines for 20 epochs on Split CIFAR-100 and 50 epochs on other benchmarks. For self-supervised pre-training, we train all baselines for 90 epochs on all benchmarks.

\begin{figure}[ht]
    \centering
    \includegraphics[width=0.95\linewidth]{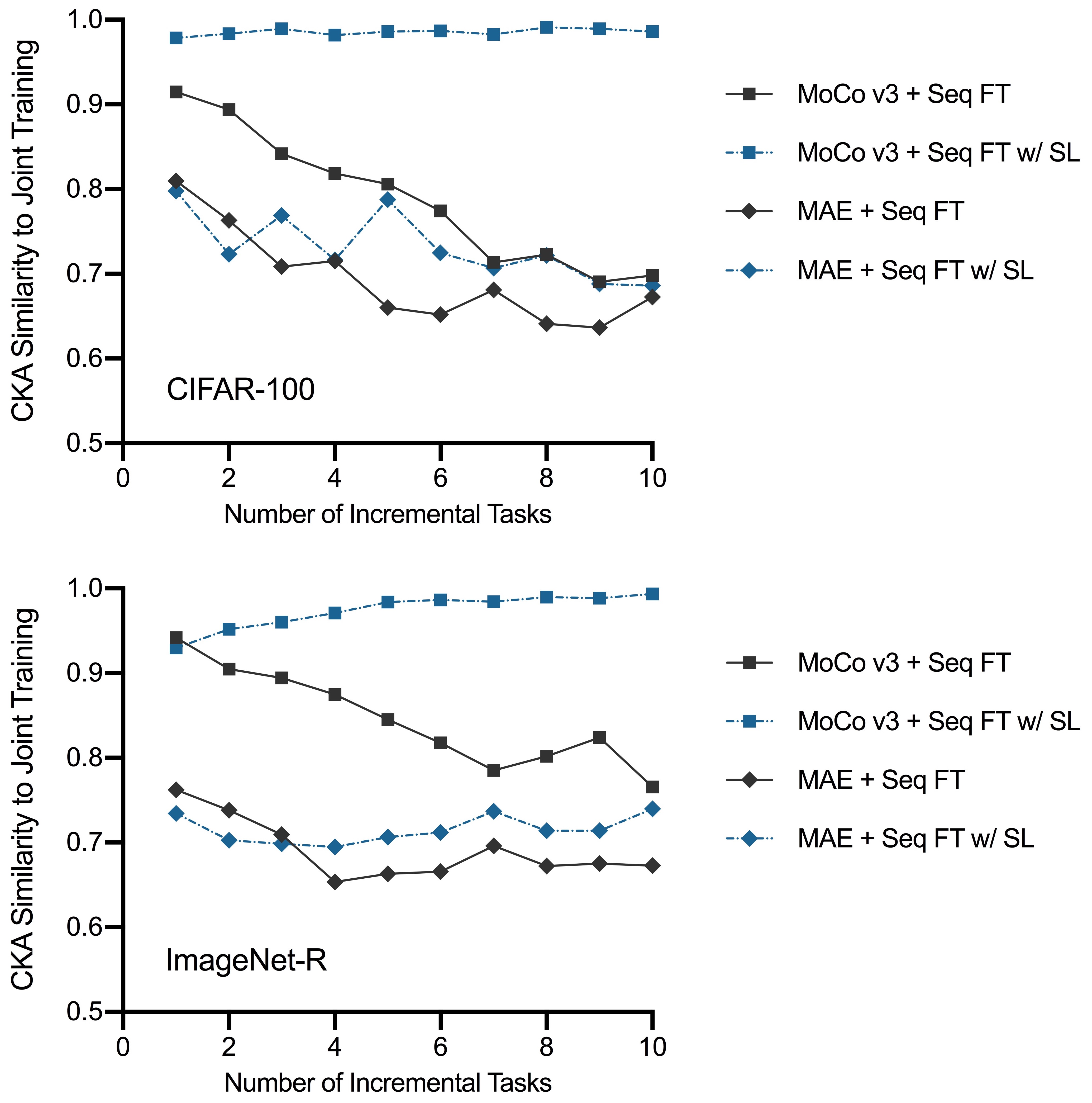}
    \caption{CKA similarity of pre-trained representations after joint training and after continual learning.
    }
    \label{ssl_cka_cl_jt}
\end{figure}

For domain-incremental setting on DomainNet benchmark, we do not use classifier alignment since the class set is fixed. We adopt the same slow learner strategy used in class-incremental setting, and the final results is obtained by averaging the classifier outputs of different domains of the same class. 

\section{Extended Analysis}
\label{sec:append_ana}
In this section, we provide extended ananlysis to support the main claims in our paper. First, In Tab.~\ref{table:learning_rate}, we provide an analysis of the impact of learning rates on CLPT using the ImageNet-21K supervised pre-training. We report Last-Acc for varying learning rates of the representation layer ranging from 0 to 0.005. A higher learning rate (0.005) for the representation layer results in even poorer performance than a fixed representation due to progressive overfitting problem. On the other hand, a too small learning rate for $\theta_{rep}$ can not improve learning on challenging  datasets like Split Cars-196, making it perform similar to the fixed representation. A moderate adjustment of learning rate for the representation layer effectively balances $M_{gen}$, $\nabla M_{plas}$ and $\nabla M_{stab}$, resolving the progressive overfitting issue and obtaining a strong continual learning performance. 

We then diagnose the behavior of using different self-supervised pre-training methods by presenting the CKA similarity of representations after each task (1) compared to the pre-trained representation in Fig.~\ref{ssl_cka_cl_pt}, and (2) compared to the joint trained representation in Fig.~\ref{ssl_cka_cl_jt}. The conclusion is inline with the conclusion from Fig. 4. For MoCo v3 pre-training, SL maintains the representation of each task close to the pre-trained representation (\textit{i.e.}, preserves $\nabla M_{gen}$), gradually increases its similarity to the  joint-trained representation (\textit{i.e.,} balances $\nabla M_{plas}$ and  $\nabla M_{stab}$). For MAE pre-training, although $\nabla M_{gen}$ is increased with SL, $\nabla M_{stab}$ from the pre-training is inadequate, resulting in a final representation that is neither close to the pre-trained nor the joint-trained one.

\begin{table}[h]
\renewcommand\arraystretch{0.95}
\caption{Ablations for CA combining with EWC and BiC. \label{table:combine}}
\resizebox{0.48\textwidth}{!}{
\begin{tabular}{lcccc}
\hline
Method & CIFAR-100 & ImageNet-R & CUB-200 & Cars-196 \\ \hline
EWC & 47.01\tiny{$\pm 0.29$} & 35.00\tiny{$\pm 0.43$} & 51.28\tiny{$\pm 2.37$} & 47.02\tiny{$\pm 3.90$} \\
EWC w/ SL &  89.30\tiny{$\pm 0.23$} & 70.27\tiny{$\pm 1.99$} & 81.62\tiny{$\pm 0.34$} & 64.50\tiny{$\pm 0.36$} \\ 
EWC w/ SL+CA &  \textbf{90.61}\tiny{$\pm 0.17$} & \textbf{71.48}\tiny{$\pm 0.31$} & \textbf{84.29}\tiny{$\pm 0.37$} & \textbf{69.61}\tiny{$\pm 0.29$} \\ 
\hline
BiC & 66.11\tiny{$\pm 1.76$} & 52.14\tiny{$\pm 1.08$} & 78.69\tiny{$\pm 1.97$} & 55.03\tiny{$\pm 3.27$} \\
BiC w/ SL & 88.45\tiny{$\pm 0.57$} & 64.89\tiny{$\pm 0.80$} & 81.91\tiny{$\pm 2.59$} &  63.10\tiny{$\pm 5.71$} \\
BiC w/ SL+CA & \textbf{91.57}\tiny{$\pm 0.13$} & \textbf{74.49}\tiny{$\pm 0.08$} & \textbf{86.82}\tiny{$\pm 0.69$} &  \textbf{73.90}\tiny{$\pm 0.38$} \\
\hline
\end{tabular}}
\end{table} 

\section{Extended Ablations}
\textbf{Combine with other methods.} 
In the main text, the efficacy of SL has been widely validated by combining it with all baseline methods. We have further validated the efficacy of CA, presenting representative non-replay and replay methods on IN21K-Sup as shown in Table~\ref{table:combine}.

\begin{table}[h]
\renewcommand\arraystretch{0.95}
\caption{Ablation studies about feature statistics in CA. \textit{shared cov.} computes covariance with momentum updates while \textit{per cls. cov.} saves covariance matrix for each class. \label{table:covs}}
\resizebox{0.48\textwidth}{!}{
\begin{tabular}{lcccc}
\hline
Method & CIFAR-100 & ImageNet-R & CUB-200 & Cars-196 \\ \hline
shared cov. & 91.06\tiny{$\pm 0.09$} & 77.38\tiny{$\pm 0.14$} & 85.64\tiny{$\pm 0.38$} & 71.39\tiny{$\pm 0.55$} \\
per cls. cov. & 91.46\tiny{$\pm 0.18$} & 78.09\tiny{$\pm 0.22$}  & 86.59\tiny{$\pm 0.29$} & 73.97\tiny{$\pm 0.22$} \\
\hline
\end{tabular}}
\end{table} 

\textbf{Influence of Feature Statistic}. In our implementation of CA, we save the covariance matrix $\Sigma_c$ of each class to estimate the per class feature distribution. As discussed in the main text, the memory efficiency can be improved by using an global $\Sigma$ for all classes by momentum updating. As shown in Tab.~\ref{table:covs}, only a slight decrease of performance is observed with this modification.

\begin{table}[h]
\renewcommand\arraystretch{0.95}
\caption{Results of our method with prompt. We compare with CODA-Prompt~\cite{smith2023coda} that also uses ImageNet-21k+ImageNet-1k pre-training.\label{table:extra_peft}}
\resizebox{0.48\textwidth}{!}{
\begin{tabular}{lccccc}
\hline
Method & \#Params & CIFAR-100 & ImageNet-R & CUB-200 & Cars-196 \\ \hline
CODA-Prompt \cite{smith2023coda} & 3.84M & 86.56\tiny{$\pm 0.77$}  & 75.25\tiny{$\pm 0.56$} &72.63\tiny{$\pm 0.76$} & 44.89\tiny{$\pm 0.61$}  \\
Ours (Prompt) & 0.22M  & \textbf{90.91}\tiny{$\pm 0.15$} & \textbf{76.46}\tiny{$\pm 0.13$}  & \textbf{81.28}\tiny{$\pm 1.91$} & \textbf{66.83}\tiny{$\pm 0.54$} \\
\hline
\end{tabular}}
\end{table} 

\textbf{SLCA++ with Prompt}. We also explore the application of our framework with the prompt tuning. We append pre-trained prompt (pre-trained on ImageNet-1K dataset) with prompt length 10 in a same way as Visual Prompt Tuning~\cite{jia2022visual}. We adopt the hybrid way of updating parameter-efficient part within the transformer. According to~\cite{gao2023unified}, prompts typically receive significantly small gradient compared to network parameters, which naturally satisfy the property of ``slow learner''. Therefore, we use a relatively large learning rate (0.02) for prompts. Compared with CODA-Prompt~\cite{smith2023coda}, our SLCA++ with prompt show a better performance, especially on fine-grained datasets, while requires much less learnable parameters. 

\end{appendices}

\vfill

\end{document}